\documentclass[a4paper,conference]{IEEEtran}
\IEEEoverridecommandlockouts
\usepackage{times}
\usepackage{epsfig}
\usepackage{graphicx}
\usepackage{amsmath}
\usepackage{amssymb}
\usepackage{amsmath}
\usepackage{algorithm}
\usepackage[noend]{algpseudocode}
\usepackage{placeins}
\usepackage{makecell}
\usepackage{boldline,multirow}
\usepackage{microtype}

\usepackage[table,xcdraw]{xcolor}
\usepackage{threeparttable}
\usepackage{subcaption}
\usepackage{colortbl}
\usepackage{multirow}
\usepackage{verbatim}

\usepackage{arydshln}
\usepackage{cite}
\usepackage[export]{adjustbox}

\usepackage{url}

\def\BibTeX{{\rm B\kern-.05em{\sc i\kern-.025em b}\kern-.08em
    T\kern-.1667em\lower.7ex\hbox{E}\kern-.125emX}}
\begin{document}

\title{Unsupervised Domain Adaptation with Multiple Domain Discriminators and Adaptive Self-Training}

\author{\IEEEauthorblockN{Teo Spadotto,
        ~Marco~Toldo,
        ~Umberto~Michieli
        ~and~Pietro~Zanuttigh}
        \IEEEauthorblockA{Department of Information Engineering, University of Padova, 35131, Padova, Italy}
        \thanks{Our work was in part supported by the Italian Minister for Education (MIUR) under the ``Departments of Excellence" initiative (Law 232/2016).}
        }

\maketitle

\begin{abstract}
Unsupervised Domain Adaptation (UDA) aims at improving the generalization capability of a model trained on a source domain to perform well on a target domain for which no labeled data is available. 
In this paper, we consider the semantic segmentation of urban scenes and we propose an approach to adapt a deep neural network trained on synthetic data to real scenes addressing the domain shift between the two different data distributions. We introduce a novel UDA framework where a standard supervised loss on labeled synthetic data is supported by an adversarial module and a self-training strategy aiming at aligning the two domain distributions. The adversarial module is driven by a couple of fully convolutional discriminators dealing with different domains: the first discriminates between ground truth and generated maps, while the second between segmentation maps coming from synthetic or real world data. The self-training module exploits the confidence estimated by the discriminators on unlabeled data to select the regions used to reinforce the learning process. Furthermore, the confidence is thresholded with an adaptive mechanism based on the per-class overall confidence.   
Experimental results prove the effectiveness of the proposed strategy in adapting a segmentation network trained on synthetic datasets like GTA5 and SYNTHIA, to real world datasets like Cityscapes and Mapillary.
\end{abstract}

\section{Introduction}
\label{sec:intro}

Deep neural networks are nowadays the most common solution to learn complex tasks from large amounts of training data. Unfortunately, they can be extremely weak in generalizing the learned knowledge to new domains, as even a small domain shift between %the two
train and test data distributions could cause serious harm to the accuracy of the network. This is particularly relevant if we aim at transferring knowledge acquired on synthetic data to the real world. Indeed, a large number of applications would benefit from training deep networks on large amounts of synthetic data. An example are autonomous driving systems, that should be able to understand the surrounding environment from visual data.
In case of road scenes, the pixel-level annotation must be manually provided for a huge amount of frames acquired by cameras mounted on moving cars, which is highly expensive, time-consuming and error-prone. %Furthermore, the car should learn to deal with uncommon critical situations, e.g., accidents, that are very difficult to be re-created in the real world scenario.
For this reason, synthetic imagery collected from graphics game engines could be extremely useful for this process \cite{Richter2016,ros2016}. 
However, despite the great realism of such scenes, a domain shift exists between computer generated data and real world images. % acquired by video cameras on cars. 
Without additional provisions, a deep network trained on synthetic images would provide low accuracy on real data.

To address this issue, we developed an UDA strategy for road driving scenes to adapt an initial learning performed on synthetic data to the real world case. 
The complete framework presented in this work is made of four components. First, a standard cross-entropy loss is employed to perform a  supervised training on synthetic data with ground truth annotations. Then, two adversarial learning schemes with a couple of fully convolutional discriminators are jointly employed. We start from the structure used in \cite{hung2018, liu2019} for semi-supervised semantic segmentation and adapted in \cite{biasetton2019, michieli2020adversarial} to Unsupervised Domain Adaptation (UDA). However, in this work we use two discriminative networks, one to differentiate between ground truth and predicted segmentations and the other to discriminate between segmentation maps coming from synthetic and real world data. 
Finally, a self-training component is introduced based on the idea that the output of the discriminator %can be used also as 
provides a measure of the reliability of the network estimations to be exploited in a self-training framework \cite{hung2018, biasetton2019}. In previous works, this module lacked two fundamental aspects: it was not class-wise adaptive and was not mutable during training. 
In other words, different classes shared the same confidence threshold to select regions for self-training, and its value was kept fixed during all the learning process.  In this work, instead, we introduce an adaptive thresholding scheme for the selection of the regions used for self-training that enforces a balanced selection for all classes and allows for variability over different training steps. Hence, the framework 
can accomplish 
both inter-class confidence flexibility and time adaptability %over the training steps.
throughout the optimization phase. 
We test our model on the task of domain adaptation for semantic segmentation of urban scenes from synthetic to real world domains. In particular, we employ the SYNTHIA and GTA5 synthetic datasets to train the supervised component, whereas we resort to real data from Cityscapes and Mapillary datasets for the unsupervised adaptation modules.

 In summary, the main contributions of this work are: (1) we introduce a novel adversarial scheme exploiting multiple domain discriminators to align the source and target domains, (2) we design a self-training module with adaptive confidence both over different classes and over different training steps, (3) we prove the effectiveness of our framework on $4$ different experimental scenarios outperforming competing approaches.

\section{Related Work}
\label{sec:related}

\noindent
\textbf{Semantic Segmentation} of images has recently witnessed substantial improvements thanks to deep learning architectures.
%The development of deep leanring has represented the key of the solution
State-of-the-art approaches for this task typically rely on an encoder-decoder structure, allowing the extraction of global semantic clues while retaining input spatial dimensionality. 
Starting from the well-known FCN architecture \cite{long2015}, many models have been proposed% have followed
, such as the PSPNet \cite{zhao2017}, DRN \cite{yu2017drn} and DeepLab in its multiple versions \cite{chen2018deeplab,chen2018encoder,chen2017rethinking} (in our work we resort to the DeepLab-v2 model \cite{chen2018deeplab} as the base segmentation network). 
The main issue affecting these highly-performing models is the high complexity, making their success strictly related to the availability of massive amounts of labelled data. For this reason, many datasets have been created (e.g., Cityscapes \cite{Cordts2016} or Mapillary \cite{neuhold2017mapillary} for %segmentation of 
urban scenes), even though the pixel-wise annotation procedure is highly expensive and time consuming. 
In light of this, %ese results, 
recent research has studied ways to exploit knowledge extracted from other sources where labels are %plentiful or 
easily accessible, and data shares as much as possible the same distribution of the target domain.

%To effectively implement this strategy, 
\textbf{Domain Adaptation}, 
especially in its unsupervised form, has drawn wide % abundant, increasingly ...
attention in the research community, as it represents a possible solution to the data shortage problem affecting several visual recognition tasks. 
Early works on domain adaptation for deep networks mainly focus on matching the statistical distribution of network embeddings, promoting domain invariant latent representations, while a task-specific loss preserves feature discriminativeness. The distribution alignment is achieved by estimating and  minimizing a measure of domain discrepancy, as %for example 
the Maximum Mean Discrepancy (MMD) \cite{TzengHZSD14, long2015learning, LongZ0J16} or a correlation distance \cite{SunFS16,SunS16}.

The introduction of adversarial learning \cite{goodfellow2014} has given rise to new adaptation techniques, which make use of an additional domain classifier %to concurrently compute a measure of domain dissimilary expressed by an adversarial loss, which is optimized in a min-max game with the task predictor.
to bridge the domain gap between source and target data distributions. 
The adversarial approach has been widely investigated in classification tasks, being applied both on the latent space, to align source and target representations \cite{ganin2015, ganin2016, tzeng2017}, and within the input space, to produce a form or target supervision provided by labeled target-like source samples \cite{LiuT16, bousmalis2017, ShrivastavaPTSW17}.    
Recently, a multitude of diverse adaptation strategies have been proposed to bridge the source-target domain discrepancy. Some examples include feature clustering for class-conditional alignment \cite{Xie2018,Deng2019}, minimization of a sliced Wasserstain distance unaffected by class imbalance \cite{Balaji2019}, feature norm matching \cite{Xu2019}, regularization provided by a joint source-target predictor \cite{Cicek2019}, entropy minimization-maximization for cross-domain feature alignment \cite{Saito2019}.

%%% DA on semantic segmentation
Unlike image classification tasks, semantic segmentation involves dense highly-structured predictions and pixel-level precision in the output space is required. 
Hence, the inherent higher complexity of network representations %
to capture both global and local semantics 
demands more sophisticated techniques to address the domain adaptation task. 
Multiple methods based on adversarial learning have been proposed, focusing on distribution alignment in the input space \cite{Zhu2017}, feature space \cite{hoffman2016}, output space \cite{tsai2018,Tsai2019} or more effectively on a combination of those \cite{hoffman2018,MurezKKRK18,sankaranarayanan2018,chen2019crdoco,toldo2020,pizzati2020domain}.
In alternative to adversarial learning, several approaches have been experimented, %explored
such as adversarial dropout \cite{Saito2018ADR, Park2018, Lee2019}, entropy minimization \cite{Chen2019, vu2019advent}, style transfer for generation of artificial target supervision \cite{Dundar18, Wu19, Choi2019} and curriculum-style learning \cite{zhang2017, lian2019constructing}.%, self-training with pseudo-labels \cite{hung2018,biasetton2019,michieli2020adversarial}.

\textbf{Self-Training} is another tool that can be exploited: 
 highly-confident predictions on unlabeled data are used to generate pseudo-labels, which are then used to train the classifier with a self-generated supervision.
While commonly employed in semi-supervised learning \cite{GrandvaletB05,hung2018}, %,liu2019}, % search for more references
self-training techniques have been recently adopted to address the domain adaptation task, as concurrently learning from source annotations and target pseudo-labels intrinsically promotes domain alignment over network embeddings \cite{biasetton2019,michieli2020adversarial,zou2018,Zou2019,lian2019constructing}.
An issue when applying self-teaching to semantic segmentation is that different classes show %have
very different distributions and %appearance
frequencies, as well as variable transferability across domains. 
Hence, a simple class-unaware confidence-based pseudo-label filtering leads %to be headed 
towards the optimization of the easy classes, whereas the harder ones are neglected, thus hindering the process of knowledge transfer to the unlabeled domain. % and thus lack of proper learning
%This in turn promotes confidence on already confident semantic categories, in a self-fueled loop that hinders the process of knowledge transfer to the unlabeled domain.
For this reason, effective self-training approaches rely on class balancing techniques, based on, for example, class weights computed \textit{a priori} over source annotations \cite{biasetton2019,michieli2020adversarial} or extrapolated from the confidence statistic of all the pixels in the target set \cite{zou2018,Zou2019}. In our approach we opt for a %lighter %and 
more effective solution, as we perform class-wise confidence threshold selection at each training step by looking at network predictions on target data in the current batch, thus avoiding assumptions on shared source-target properties, or slow computations involving the analysis of the whole employed dataset.

\section{Proposed Domain Adaptation Strategy}
\label{sec:method}

\begin{figure*}[htbp]
\centering
\includegraphics[width=\linewidth]{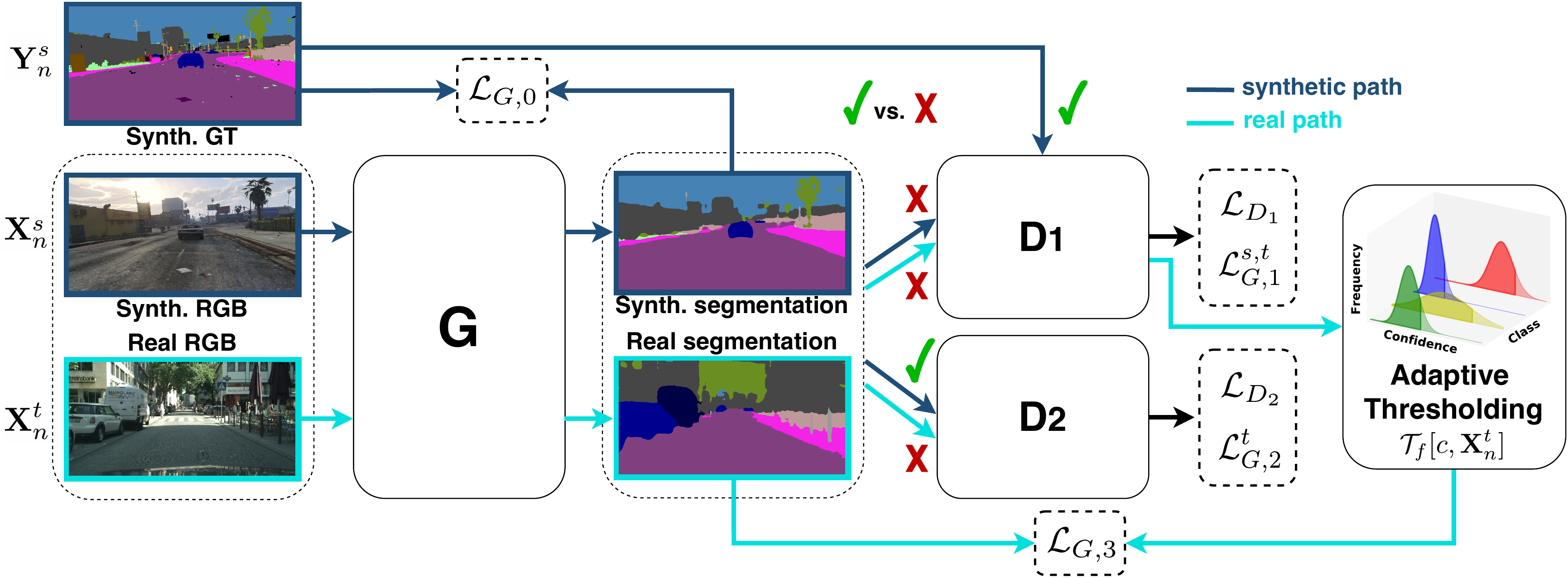}
\caption{Architecture of the proposed approach (\textit{best viewed in colors}). The semantic segmentation network $G$ is trained with the combination of $4$ losses: a supervised cross entropy on source data $\mathcal{L}_{G,0}$, a double adversarial framework $\mathcal{L}^{s,t}_{G,1}$ and $\mathcal{L}_{G,2}^t$, and a self-training module $\mathcal{L}_{G,3}$ with class-wise and time-varying adaptive thresholding mask $\mathcal{T}_f$.}
\label{fig:architecture}
\end{figure*}

In this section the complete architecture of the proposed approach is %presented and 
discussed. An overview of the method is given in Figure~\ref{fig:architecture}. We denote with $G$ the semantic segmentation network that we want to adapt from supervised synthetic data to unlabeled real data. $G$ takes the role of the generator in our adversarial setup.
The optimization of the segmentation network is driven by the minimization of a multi-target objective involving four loss functions. In particular, to guide the adaptation of $G$ to unlabeled real data we employ two discriminative networks and a self-training module. %The training of the framework is guided by the combination of 4 loss functions.

Let us denote with $\mathbf{X}^s_n$  the generic $n$-th image in the source (synthetic) domain and with $\mathbf{Y}^s_n$ the corresponding ground truth segmentation, while $\mathbf{X}^t_n$ is the $n$-th real world sample (for which ground truth is not available during training). 
%A supplementary discriminator network $D$ is used to evaluate the reliability of $G$'s output. T
%In this section, we detail the CNN architectures and the training procedure implementing the unsupervised domain adaptation. 
The first component of our approach, $\mathcal{L}_{G,0}$, is a standard  cross-entropy loss working on labeled synthetic data.

The second and the third loss functions, minimized during the training of $G$, are relative to the adversarial training with the couple of discriminator networks. The first discriminator module $D_1$ is trained to distinguish between ground truth and generated maps (the latter either coming from synthetic or real images). The peculiarity of this  network is that it is a fully convolutional model and it produces a per-pixel confidence estimation, differently from traditional adversarial frameworks where the discriminator outputs a single binary value for the whole input image.
The key idea is that 
the discriminative action provides a measure of discrepancy between the statistic of source annotations and both source and target prediction maps, and its optimization leads to an indirect yet effective cross-domain distribution alignment \cite{hung2018,biasetton2019,michieli2020adversarial}. %over the segmentation output space, thus 

The discriminator network $D_1$ is made of a stack of $5$ convolutional layers each using $4\times 4$ kernels with a stride of $2$ and Leaky ReLU activation function. The number of filters (from first to last layer) is $64$, $64$, $128$, $128$, $1$ and the cascade is followed by a bilinear upsampling to match the original input image resolution. 
The network $D_1$ is  trained to minimize the loss $\mathcal{L}_{D_1}$ that is a standard cross-entropy loss between $D_1$'s output and the one-hot encoding indicating whether the input segmentation map is the ground truth (class $1$) or it is produced by $G$ (class $0$), i.e.: % $\mathcal{L}_{D_1}$ is then defined as:
\begin{equation}
%\begin{split}
\mathcal{L}_{D_1} \!=\! -\!\! \sum_{p\in \mathbf{X}^{s,t}_n}\!\! \log(1\!-\!D_1(G(\mathbf{X}^{s,t}_n))^{( p )}) + \log(D_1(\mathbf{Y}^{s}_n)^{( p )})
%\end{split}
\label{eq:L_D1}
\end{equation}
where $p$ is a generic pixel in the image.
Meanwhile, $G$ %, playing against $D_1$ as in a standard adversarial framework,
 is forced to produce segmentation maps that resemble ground-truth annotations when %processed
inspected by $D_1$:
%On a generic image $\mathbf{X}^{s,t}_n$ this objective can be translated into the following loss function:
%
\begin{equation}
\mathcal{L}^{s,t}_{G,1} = - \sum_{p\in \mathbf{X}^{s,t}_n}   \log(D_1(G(\mathbf{X}^{s,t}_n))^{(p)})
\label{eq:L_G2}
\end{equation}
The second discriminator, $D_2$, is trained to differentiate between segmentation maps coming from synthetic or real-world data. Differently from $D_1$, $D_2$ is always fed with the generator output, i.e. with source or target images  segmented by $G$, whereas no ground-truth information is employed. $D_2$ has the same structure as $D_1$ except that its number of channels has been reduced to $48$, $48$, $96$, $96$, $1$, as its adaptation objective is complementary to the one of $D_1$ and requires less computational complexity to be accomplished.
The adversarial loss for $D_2$ can be expressed as:
%Novel contribution to further align,... idea is that...
%
%
\begin{equation}
%\begin{split}
\mathcal{L}_{D_2} \!=\! -\!\!\!\!\sum_{p\in \mathbf{X}^{s,t}_n}\!\!\! \! \log(1\!-\!D_2(G(\mathbf{X}^{t}_n))^{(p)}) + \log( \! D_2(G(\mathbf{X}^{s}_n))^{( p )})
%\end{split}
\label{eq:L_D2}
\end{equation}
which is a standard cross-entropy loss similar to $\mathcal{L}_{D_1}$. Here the objective is formulated between $D_2$'s output and the one-hot encoding marking with $0$ and $1$ the segmentation maps produced by $G$ from respectively target and source images.
The role of $G$ in the second adversarial competition is to fool $D_2$, by computing sufficiently realistic target prediction maps resembling the source domain ones. % with sufficient accuracy.
Hence, similarly to Eq. \ref{eq:L_G2}, the adversarial loss for $G$ related to $D_2$ is of the form:
\begin{equation}
\mathcal{L}^{t}_{G,2} = - \sum_{p\in \mathbf{X}^{t}_n}   \log(D_2(G(\mathbf{X}^{t}_n))^{(p)})
\label{eq:L_G2_D2}
\end{equation}
The second adversarial framework is specifically focused on the adaptation of target network representations, in that $G$ is trained to produce target segmentation maps which are close to source ones from a statistical point of view. Thus it directly tackles the domain gap causing the performance drop on the target set.
Source-target alignment also happens within the first adversarial scenario %in an indirect form,
as a side effect. Indeed, $G$'s output is forced to be distributed as ground-truth labels for both source and target inputs, but no actual cross-domain adaptation of network embeddings is directly performed.

%The effectiveness of the additional adversarial adaptation is proved by numerical results shown in section \ref{}.
%

The last module in our adaptation framework implements a self-training strategy. As shown by recent works \cite{hung2018,biasetton2019,michieli2020adversarial}, the output of the discriminator $D_1$ can be interpreted  as  a measure of confidence for predicted target pixels, since $D_1$ should identify target predictions that deviate from source ground-truth statistic. Hence, the output of $D_1$ can be exploited to select the reliable predictions in the target segmentation maps. The selected output samples are then converted into one-hot encoded data, which can be used as a self-taught ground-truth to train $G$ directly on unlabeled target samples. The self-training objective is expressed by the following loss: 
\begin{equation}
\mathcal{L}_{G,3} \!= \!-\! \!\! \! \sum_{p\in \mathbf{X}_n^t} \!
\sum_{c \in \mathcal{C}}
%\! D_{\!TH}(\mathbf{X}_n^t)^{(p)} \! \cdot
\mathcal{M}_f^{(p)} \cdot
W_c^s \cdot \, 
\!\mathbf{\hat{Y}}_n^{(p)}[c] \cdot
\log\!\big(G(\mathbf{X}_n^t)^{(p)}[c]\big)
\end{equation}
where $\mathbf{\hat{Y}}_n$ is the one-hot encoded ground truth derived from the per-class argmax of the generated probability map $G(\mathbf{X}_n^t)$, $c$ is a specific class belonging to the set of possible classes $\mathcal{C}$ and $W_c^s$ is a weighting function proportional to the class frequency on the source domain as introduced in \cite{biasetton2019,michieli2020adversarial}. 
%$\mathcal{M}_f^{(p)}$ is the adaptive thresholding mask for the selection of the regions used for self-training, defined as:
$\mathcal{M}_f^{(p)}[c, \mathbf{X}_n^t]$ (for ease of notation we drop its dependencies on % avoid to explicitly report its dependency on
class and training sample in all the equations) is the adaptive thresholding mask for the selection of the regions used for self-training, defined as:
\begin{equation}
\mathcal{M}_f^{(p)}\!= \!
\begin{cases} 
1 & \mathrm{if}  \left( D_1(G(\mathbf{X}_n^t))^{(p)}\!>\!\mathcal{T}_f[c, \mathbf{X}_n^t] \right) \, \land \, \left( \! \mathbf{\hat{Y}}_n^{(p)}[c] =\! 1 \!\right)\\
0 &  \mathrm{otherwise} \\
\end{cases}
\label{eq:mask}
\end{equation}
Differently from previous works \cite{biasetton2019,michieli2020adversarial,Zou2019,lian2019constructing}, in which the confidence threshold was computed before-hand and kept fixed throughout the training, here we propose to have both a class-level and a training-stage adaptation of the threshold. Indeed, different classes typically have different confidence values which may also vary during training. The class-wise confidence threshold selection for a generic training step and a generic class $c$ is formulated as:
\begin{equation}
\label{eq:thresholds}
\mathcal{T}_f[c, \mathbf{X}_n^t] =   \mathcal{Q}_f ( D_1(G(\mathbf{X}_n^t)[c]) )
\end{equation}
where $\mathcal{Q}_f$ represents the $f$-th percentile. The best results are obtained with $f$ in the range of $75-80\%$  to enable self-training only in regions with high confidence on target data. We compute $\mathcal{T}_f$ for every class at each training step by looking at network predictions on target data in the current batch. This makes the model adaptive both to the statistic of the various classes and to the different training phases.
Finally, we can compute the overall loss function for $G$ with a weighted average of the four individual losses, i.e.:
\begin{equation}
\label{eq:L_full}
\mathcal{L}_{full} = \mathcal{L}_{G,0} + w_{1}^{s,t}\mathcal{L}_{G,1}^{s,t}+ w_{2}^{t}\mathcal{L}_{G,2}^{t} + w_{3} \mathcal{L}_{G,3}
\end{equation}
with weighting parameters $w^{s,t}_1$, $w^t_2$ and $w_3$ empirically set.% discovered. 
%where weighting parameters $w^{s,t}_1$, $w^t_2$  and $w_3$ are empirically discovered.
%
%
%

\section{Experimental Results}
\label{sec:results}

\subsection{Datasets}
We evaluate our framework on some publicly available and widely used datasets. 
The supervised synthetic training is performed on the GTA5 \cite{Richter2016} and  SYNTHIA \cite{ros2016} datasets. The real world datasets for unsupervised adaptation and for results evaluation are Cityscapes \cite{Cordts2016} and Mapillary \cite{neuhold2017mapillary}. 
The evaluation scenario is the same of competing approaches as \cite{hoffman2016, sankaranarayanan2018, zhang2017, biasetton2019, michieli2020adversarial} to allow for a fair comparison. 
We provide the segmentation network with images of $750\times 375 \mathrm{px}$ during training for memory constraints, while testing is done at the original resolution.

\textbf{GTA5} \cite{Richter2016} contains $24966$ synthetic images acquired through the video game \textit{Grand Theft Auto 5} and labeled into $19$ classes compatible with real datasets. The images are taken from the car perspective in American-style virtual cities and are characterized by a high level of quality and realism.
We used $23966$ images for supervised training while $1000$ have been taken out for validation purposes. %There are $19$ semantic classes which are compatible with the ones of Cityscapes. 

\textbf{SYNTHIA} \cite{ros2016} contains  $9400$ synthetic images rendered with an ad-hoc engine (we used the \textit{SYNTHIA-RAND-CITYSCAPES} subset). 
Frames come from virtual European-style towns with a large variability of scenes  under various light and weather conditions.
On the other hand, the visual quality is lower than that of GTA5. 
For the evaluation on real datasets, only $16$ overlapping classes are taken into consideration.
We used $9300$ images for supervised training and $100$ for validation purposes.

\textbf{Cityscapes} \cite{Cordts2016} contains $2975$ images %$2048 \times 1024\mathrm{px}$
 acquired in $50$ European cities.  
The original training set (without labels) has been used for unsupervised adaptation, while the $500$ images in the original validation set have been used as a test set (as done by competing approaches since the test set labels are not publicly available).

\textbf{Mapillary} \cite{neuhold2017mapillary} contains $20000$ high resolution images, whose %strong
diversity in terms of semantic categories and acquisition settings ensure %sufficient high
improved data expressiveness% and completeness to learn the complex task of semantic segmentation.
, essential to address the semantic segmentation task.
We employed all $18000$ images from the training set (without labels) when performing unsupervised adaptation in the training stage, while we resorted to the validation set of $2000$ images for testing, as done by  competing approaches. Following \cite{kim2018attribute}, we perform class matching to the Cityscapes dataset.

\subsection{Training Details}
The approach is agnostic to the deep learning architecture used for $G$: in principle any semantic segmentation network can be used, in our case we employ the well known Deeplab v2 model \cite{chen2018deeplab} with the ResNet-101 backbone.
The  weights are pre-trained \cite{nekrasov} on the MSCOCO dataset \cite{lin2014microsoft}.

The proposed deep learning scheme is implemented in TensorFlow and the code will be available soon. 
The  generator network $G$ is trained as proposed in \cite{chen2018deeplab} using the Stochastic Gradient Descent (SGD)  optimizer with momentum set to $0.9$ and weight decay to $10^{-4}$. 
The discriminators $D_1$ and $D_2$ are trained using the Adam optimizer. The learning rate employed for both $G$ and $D_1$ starts from $10^{-4}$ and is decreased up to $10^{-6}$ by means of a polynomial decay with power $0.9$. 
As for $D_2$, the base learning rate is set to $10^{-4}$ and $5\times10^{-4}$ for respectively the SYNTHIA and GTA5 adaptation scenarios. %, and it is kept fixed throughout the training phase. 
The weighting parameters %of $\mathcal{L}_{full}$ 
are empirically set as $w^{s}_1=10^{-2}$, $w^{t}_1=10^{-3}$ and $w^t_2=10^{-2}$ independently of the source and target datasets. The self-training parameter $w_3$ is fixed to $5\times10^{-2}$ and $10^{-1}$ for respectively SYNTHIA and GTA5 cases, as the inferior realism of the SYNTHIA dataset suggests a more cautios usage of the self-training module.
We train the model for $20K$ iterations %r The number of training steps (and also decay steps) was set to , that 
on a NVIDIA RTX 2080 Ti GPU. The longest training inside this work, i.e., the one with all the loss components enabled, takes about $6$ hours to complete.

\begin{table*}[tbp]
\setlength{\tabcolsep}{2.75pt}
\renewcommand{\arraystretch}{1.1}

\centering
% GTA --> CITYSCAPES
\begin{tabular}{c|c|c|c|ccccccccccccccccccc|c|}
 \cline{2-24}
& & & Method & \rotatebox{90}{road} &  \rotatebox{90}{sidewalk} &  \rotatebox{90}{building} &  \rotatebox{90}{wall} &\rotatebox{90}{fence} & \rotatebox{90}{pole} & \rotatebox{90}{t light} 
  &\rotatebox{90}{t sign} & \rotatebox{90}{veg} & \rotatebox{90}{terrain} & \rotatebox{90}{sky} & \rotatebox{90}{person}& \rotatebox{90}{rider} & \rotatebox{90}{car} 
  & \rotatebox{90}{truck} & \rotatebox{90}{bus} & \rotatebox{90}{train} &  \rotatebox{90}{mbike} & \rotatebox{90}{bike} & \rotatebox{90}{mean} \\
 \cline{2-24}
\multirow{7}{*}{\vspace{0ex}a)} & \multirow{14}{*}{\rotatebox{90}{\hspace{0ex}To Cityscapes}} & \multirow{7}{*}{\rotatebox{90}{\scriptsize From GTA5\hspace{0ex}}} 
& Supervised ($\mathcal{L}_{G,0}$ only) & 49.3 & 24.4 & 56.4 &  6.5 & \textbf{19.6} & 25.6 & \textbf{23.6} & 10.1 & 82.7 & 28.5 & 69.9 & \textbf{55.5} &  4.9 & 80.9 & 18.0 & 33.0 &  1.2 & 15.1 &  0.1 & 31.9 \\   
& & & Hoffman et al. \cite{hoffman2016} & 70.4 &  32.4 & 62.1 & 14.9 & 5.4 & 10.9 & 14.2 &  2.7 &   79.2 &  21.3 & 64.6 & 44.1 &  4.2&  70.4 & 8.0 & 7.3 & 0.0 & 3.5 & 0.0 & 27.1\\
& & & Hung et al. \cite{hung2018} & \textbf{81.7} & 0.3 & 68.4 & 4.5 & 2.7 & 8.5 & 0.6 & 0.0 & 82.7 & 21.5 & 67.9 & 40.0 & 3.3 & 80.7 & \textbf{34.2} & \textbf{45.9 }& 0.2 & 8.7 & 0.0 & 29.0\\
& & & Zhang et al. \cite{zhang2017} & 74.9 & 22.0 & \textbf{71.7} & 6.0 & 11.9 & 8.4 & 16.3 & \textbf{11.1} & 75.7 & 13.3 & 66.5 & 38.0 & \textbf{9.3} & 55.2 & 18.8 & 18.9 & 0.0 & \textbf{16.8} & \textbf{14.6} & 28.9 \\
& & & Biasetton et al. \cite{biasetton2019} & 54.9 & 23.8 & 50.9 & 16.2 & 11.2 & 20.0 & 3.2 & 0.0 & 79.7 & 31.6 & 64.9 & 52.5 & 7.9 & 79.5 & 27.2 & 41.8 & 0.5 & 10.7 & 1.3 & 30.4\\
& & & Michieli et al. \cite{michieli2020adversarial} & 81.0 & 19.6 & 65.8 & \textbf{20.7} & 2.9 & 20.9 & 6.6 & 0.2 & 82.4 & 33.0 & 68.2 & 54.9 & 6.2 & 80.3 & 28.1 & 41.6 & \textbf{2.4} & 8.5 & 0.0 & 33.3 \\
& & & \textbf{Ours} & 77.7 & \textbf{35.9} & 67.2 & 18.9 & 12.1 & \textbf{26.2} & 15.9 &  5.9 & \textbf{83.7} & \textbf{33.3} & \textbf{72.7} & 53.9 &  4.2 & \textbf{82.6} & 21.5 & 41.1 &  0.1 & 13.9 &  0.0 & \textbf{35.1} \\
 \cline{3-24}

% SYNTHIA --> CITYSCAPES
\multirow{7}{*}{\vspace{0ex}b)} & & \multirow{7}{*}{\rotatebox{90}{\scriptsize  From SYNTHIA\hspace{0ex}}} 
& Supervised ($\mathcal{L}_{G,0}$ only) & 17.9 & 24.2 & 38.6 &  \textbf{5.0} &  0.0 & 28.7 &  0.0 &  4.5 & 79.3 & - & 80.8 & \textbf{54.0} &  \textbf{8.9} & 75.7 & - & 35.4 & - & 4.2 & 3.9 & 28.8 \\
& & & Hoffman et al. \cite{hoffman2016} & 11.5 & 19.6 & 30.8 & 4.4 & 0.0 & 20.3 & 0.1 & \textbf{11.7} & 42.3  & - & 68.7 & 51.2 & 3.8 & 54.0  & - &  3.2  & - & 0.2 & 0.6 & 20.1  \\
& & & Hung et al. \cite{hung2018} & 72.5 & 0.0 & 63.8 & 0.0 & 0.0 & 16.3 & 0.0 & 0.5 & \textbf{84.7}  & - & 76.9 & 45.3 & 1.5 & 77.6  & - & 31.3 & - &  0.0 & 0.1 & 29.4 \\
& & & Zhang et al. \cite{zhang2017} & 65.2 & 26.1 & 74.9 & 0.1 & \textbf{0.5} & 10.7 & \textbf{3.7} & 3.0 & 76.1  & - & 70.6 & 47.1 & 8.2 & 43.2  & - & 20.7 & - & 0.7 & \textbf{13.1} & 29.0 \\
& & & Biasetton et al. \cite{biasetton2019} & 78.4 & 0.1 & 73.2 & 0.0 & 0.0 & 16.9 & 0.0 & 0.2 & 84.3 &  - & 78.8 & 46.0 & 0.3 & 74.9 &  - & 30.8 & - & 0.0 & 0.1 & 30.2  \\
& & & Michieli et al. \cite{michieli2020adversarial} & \textbf{80.7} & 0.3 & \textbf{75.0} & 0.0 & 0.0 & 19.5 & 0.0 & 0.4 & 84.0  & - & 79.4 & 46.6 & 0.8 & 80.8  & - & 32.8  & - & 0.5 & 0.5 & 31.3 \\
& & & \textbf{Ours} & 72.0 & \textbf{26.6} & 66.1 &  1.8 &  0.0 & \textbf{30.2} &  0.0 &  4.3 & 81.4  & - & \textbf{82.2} & 51.7 &  4.0 & \textbf{82.2} &  - & \textbf{37.9} & - &  \textbf{7.7} &  5.9 & \textbf{34.6} \\
\clineB{2-24}{3}

% GTA5 --> MAPILLARY
\multirow{5}{*}{\vspace{0ex}c)} & \multirow{10}{*}{\rotatebox{90}{\hspace{0ex}To Mapillary}}  & \multirow{5}{*}{\rotatebox{90}{\scriptsize  From GTA5\hspace{0ex}}} & 
Supervised ($\mathcal{L}_{G,0}$ only) & 69.8 & 31.8 & 58.8 & 14.6 & 22.3 & 28.3 & \textbf{31.8} & \textbf{28.8} & 70.0 & 24.4 & 72.4 & \textbf{60.4} & 16.8 & 80.6 & 36.6 & 34.3 & \textbf{10.2} & \textbf{26.2} &  \textbf{0.2} &  37.8 \\
& & & Hung et al. \cite{hung2018} & 78.2 & 29.7 & 68.7 & 10.0 & 6.7 & 17.5 & 0.0 & 0.0 & 76.4 & 35.2 & \textbf{95.6} & 53.8 & 13.8 & 77.5 & 34.3 & 30.2 & 5.0 & 21.8 & 0.0 & 34.4 \\
& & & Biasetton et al. \cite{biasetton2019} & 71.4 & 25.0 & 62.0 & 20.4 & 17.6 & 26.8 & 5.9 & 0.8 & 64.6 & 24.6 & 86.5 & 58.3 & 14.7 & 80.0 & 39.3 & 42.2 & 5.5 & 22.3 & 0.1 & 35.2 \\
& & & Michieli et al. \cite{michieli2020adversarial} & 79.9 & 28.0 & 73.4 & \textbf{23.0} & 29.5 & 20.9 & 1.1 & 0.0 & \textbf{79.5} & \textbf{39.6} & 95.0 & 57.6 & 9.0 & 80.6 & 41.5 & 40.1 & 7.4 & 24.8 & 0.1 & 38.5 \\
& & & \textbf{Ours} & \textbf{80.0} & \textbf{43.3} & \textbf{75.4} & 19.4 & \textbf{29.7} & \textbf{29.6} & 23.3 & 16.2 & 78.5 & 33.5 & 93.7 & 59.0 & \textbf{20.3} & \textbf{82.2} & \textbf{44.5} & \textbf{43.4} &  2.5 & 22.1 &  0.0 & \textbf{41.9} \\
 \cline{3-24}

%SYNTHIA --> MAPILLARY

\multirow{5}{*}{\vspace{0ex}d)} & & \multirow{5}{*}{\scriptsize \rotatebox{90}{From SYNTHIA\hspace{0ex}}}  & 
Supervised ($\mathcal{L}_{G,0}$ only) & 25.4 & 22.0 & 56.4 &  \textbf{6.9} &  \textbf{0.1} & 29.4 &  0.0 &  \textbf{2.8} & 72.8  & - & 92.1 & 53.7 & \textbf{16.1} & 75.1  & - & \textbf{30.8}  & - &  8.6 &  5.8 &  31.1 \\
& & & Hung et al. \cite{hung2018} & 36.8 & 20.1 & 53.9 & 0.0 & 0.0 & 23.7 & 0.0 & 0.0 & 73.9  & - & \textbf{95.6} & 43.4 & 0.1 & 64.6  & - & 19.0 & -  & 0.4 & 0.5 & 27.0 \\
& & & Biasetton et al. \cite{biasetton2019} & 16.4 & 19.1 & 42.2 & 2.7 & 0.0 & \textbf{33.1} & 0.0 & 1.3 & 76.5  & - & 88.0 & 50.4 & 10.9 & 69.9 & -  & 25.5 & -  & 6.1 & 9.2 & 28.2 \\
& & & Michieli et al. \cite{michieli2020adversarial} & 57.6 & 18.3 & 62.1 & 0.4 & 0.0 & 23.7 & 0.0 & 0.0 & \textbf{79.4} & -  & 94.8 & 52.4 & 9.2 & 74.2  & - & 28.3  & - & 4.0 & 6.9 & 32.0  \\
& & & \textbf{Ours} & \textbf{59.0} & \textbf{28.4} & \textbf{68.6} &  0.7 &  0.0 & 29.8 &  0.0 &  1.8 & 77.5  & - & 94.9 & \textbf{54.6} & 12.6 & \textbf{76.8}  & - & 28.0  & - & \textbf{11.2} & \textbf{14.7} &  \textbf{34.9}  \\ \cline{2-24}
\end{tabular}

\caption{Per-class and mean IoU on the four considered UDA scenarios. The approaches have been trained in a supervised way on the synthetic dataset and the unsupervised domain adaptation has been performed using the respective real world training set. The results are reported on the real world validation sets.}
\label{tab:quantitative}
\end{table*}

\subsection{Evaluation on the Cityscapes Dataset}

The first scenario we consider for evaluation purposes comprises the adaptation to the Cityscapes dataset from both SYNTHIA and GTA5. As done by competing approaches \cite{hoffman2016, chen2018road, tsai2018}, the numerical performance is expressed in terms of mean Intersection over Union (mIoU) between predicted maps and ground-truth labels over the Cityscapes validation set. The per-class mIoU results of the evaluations are displayed in Table \ref{tab:quantitative}a) and \ref{tab:quantitative}b).  
We denote as \textit{supervised} the na\"ive approach relying only on source supervision and no form of adaptation, while the numerical performance of our method as a whole is reported in the last row of each section. 
%

%To start, we consider the GTA5 dataset as source domain.
Using the GTA5 as source domain, the simple supervised approach achieves a final mIoU of $31.8\%$ on the target dataset. 
The introduction of the multi-domain adversarial learning scheme and adaptive self-training module leads to a performance increment of $3.3\%$, boosting the mIoU up to $35.1\%$. % when the complete adaptation framework in employed. 
Moreover, the improvement is shared by the majority of the semantic classes.
Some categories characterized by semantic similarity between them (such as \textit{road}, \textit{sidewalk} and \textit{terrain}) or with appearance discrepancy across source and target domains (such as \textit{car}, \textit{truck} and \textit{bus}) seem to highly benefit from the adaptation, proving that our method succeeded in bridging the domain gap.
%showing how our approach is able to bridge the domain gap and to successfully transfer acquired knowledge to the target domain. 
In Table \ref{tab:quantitative} we also report results achieved by some competing approaches. It can be noticed that our strategy outperforms all of them, including those relying on simpler self-training and adversarial learning schemes (i.e., \cite{hung2018, biasetton2019, michieli2020adversarial}), demonstrating the efficacy of the multi-domain discrimination and of the adaptive thresholding techniques we introduced. In Figure \ref{fig:qual_res}a) we show some sample segmented images from the Cityscapes validation set. We can appreciate that our approach leads to a more precise detection of several semantic entities found in the input image. For example, the \textit{road} and \textit{sidewalk} classes are subject to a more accurate recognition, which supports the numerical results. % as just discussed.

Section (b) of Table \ref{tab:quantitative} reports the results of the adaptation from the SYNTHIA dataset. As for the GTA5 case, our adaptation strategy represents a considerable improvement over the supervised training on the source dataset. The mIoU achieved without adaptation ($28.8\%$) is pushed up to $34.6\%$ by our framework. 
The increment of almost $6\%$ is even higher than %when the GTA5 dataset is employed
the one achieved with the more realistic GTA5 dataset as source domain, and this proves the effectiveness of the adaptation modules we developed also in a challenging environment with a larger statistical gap across domains. 
For example, while texture discrepancy between Cityscapes and SYNTHIA causes the road on real-world scenes to be hardly recognized by the segmentation network in lack of adaptation, our strategy successfully provides the predictor with road detection capabilities. 
Figure \ref{fig:qual_res}a) includes some qualitative results. 
Again, we can observe the improved semantic understanding and detection accuracy on classes such as \textit{road} and \textit{sidewalk} with respect to the baseline, as well as with respect to a competing approach based on adversarial and self-training techniques \cite{biasetton2019}.

\newcommand{\imgsize}{32.85mm}
\begin{figure*}[htbp]
\centering
\begin{subfigure}[htbp]{\textwidth}
%\hspace{0.05mm}
\hspace{0.22cm}
\resizebox{0.965\textwidth}{!}{%
\begin{tabular}{cccccccccc}
\cellcolor[HTML]{804080}{\color[HTML]{FFFFFF} \textbf{road}} & \cellcolor[HTML]{F423E8}\textbf{sidewalk} & \cellcolor[HTML]{464646}{\color[HTML]{FFFFFF} \textbf{building}} & \cellcolor[HTML]{66669C}{\color[HTML]{FFFFFF} \textbf{wall}} & \cellcolor[HTML]{BE9999}\textbf{fence} & \cellcolor[HTML]{999999}\textbf{pole} & \cellcolor[HTML]{FAAA1E}\textbf{traffic light} & \cellcolor[HTML]{DCDC00}\textbf{traffic sign} &\cellcolor[HTML]{6B8E23} \textbf{vegetation} & \cellcolor[HTML]{98FB98}\textbf{terrain} \\ \cline{10-10} 
\cellcolor[HTML]{4682B4}\textbf{sky} & \cellcolor[HTML]{DC143C}{\color[HTML]{FFFFFF} \textbf{person}} & \cellcolor[HTML]{FF0000}{\color[HTML]{FFFFFF} \textbf{rider}} & \cellcolor[HTML]{00008E}{\color[HTML]{FFFFFF} \textbf{car}} & \cellcolor[HTML]{000046}{\color[HTML]{FFFFFF} \textbf{truck}} & \cellcolor[HTML]{003C64}{\color[HTML]{FFFFFF} \textbf{bus}} & \cellcolor[HTML]{005064}{\color[HTML]{FFFFFF} \textbf{train}} & \cellcolor[HTML]{0000E6}{\color[HTML]{FFFFFF} \textbf{motorcycle}} & \multicolumn{1}{c|}{\cellcolor[HTML]{770B20}{\color[HTML]{FFFFFF} \textbf{bicycle}}} & \multicolumn{1}{c|}{\textbf{unlabeled}} \\ \cline{10-10} 
\end{tabular}%
}
\vspace{0.1cm}
\end{subfigure}
\setlength{\tabcolsep}{1pt} % Default value: 6pt
\centering
\begin{subfigure}[htbp]{2\textwidth}
\begin{tabular}{c|c|c|ccccc}
\cline{2-3}
  
  \multirow{4}{*}{\vspace{-23ex}a)} & \multirow{4}{*}{\rotatebox{90}{\hspace{-18ex}To Cityscapes}} & \multirow{2}{*}{\rotatebox{90}{From GTA5\hspace{+1ex}}} &
  \includegraphics[width=\imgsize , valign=c]{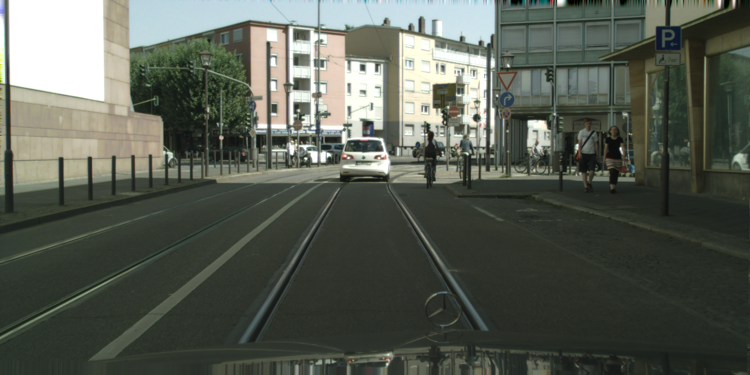} &
   \includegraphics[width=\imgsize , valign=c]{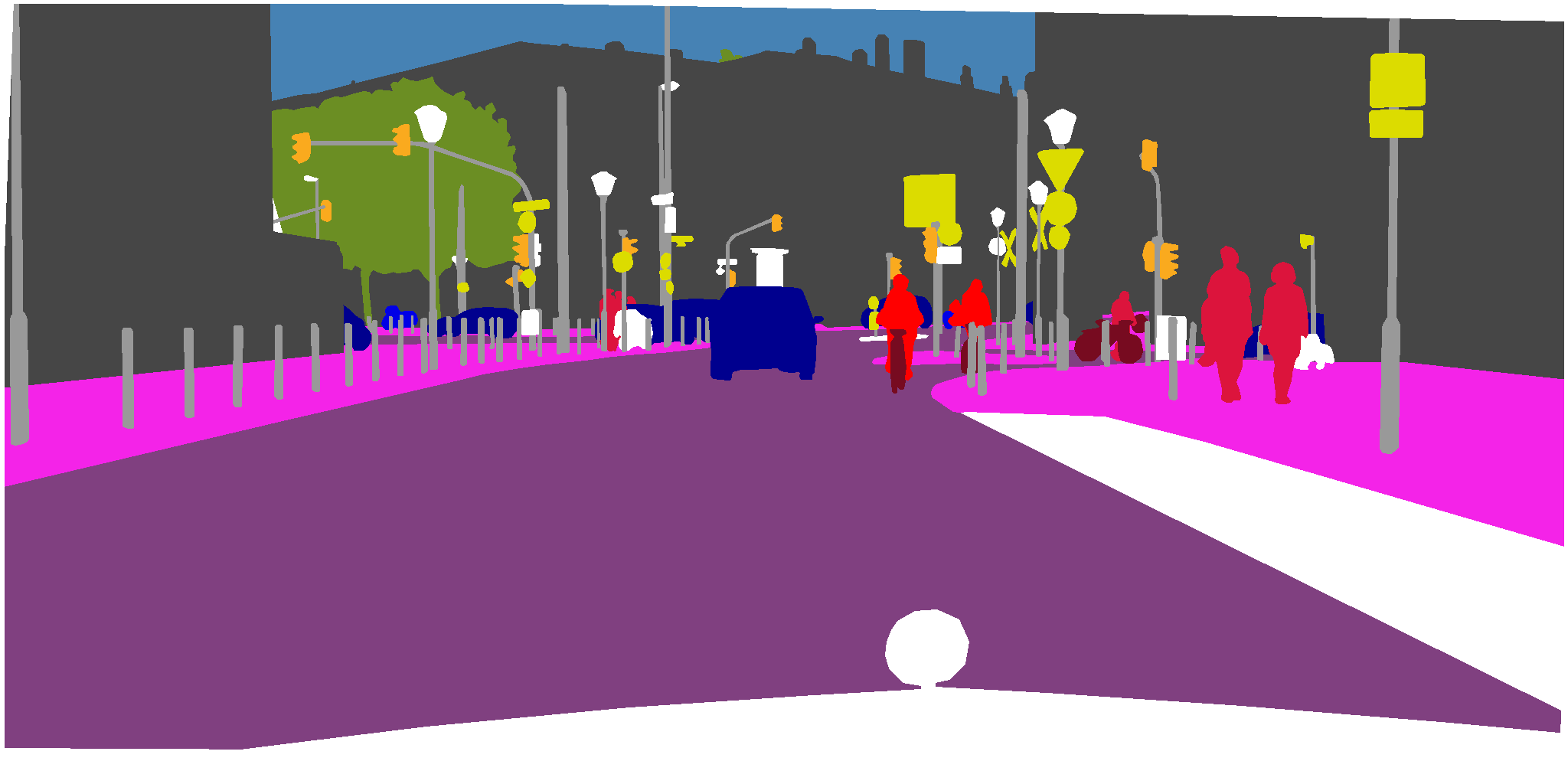} & 
   \includegraphics[width=\imgsize , valign=c]{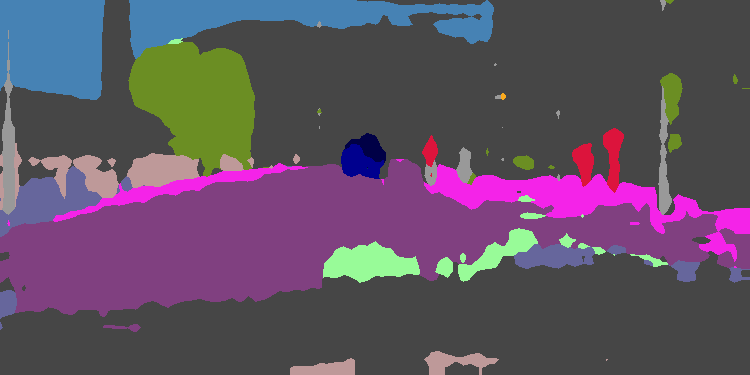} & 
   \includegraphics[width=\imgsize , valign=c]{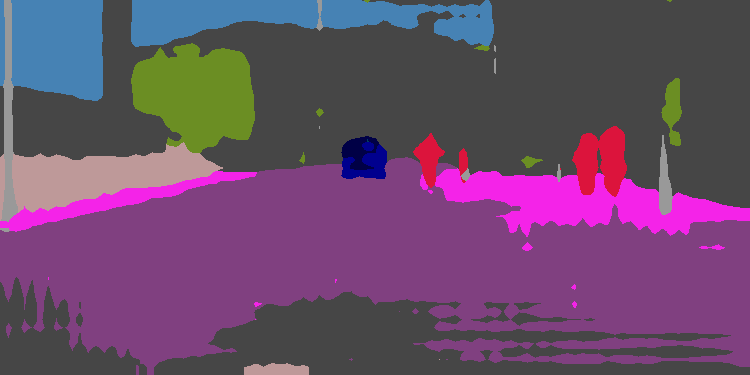} & 
   \includegraphics[width=\imgsize , valign=c]{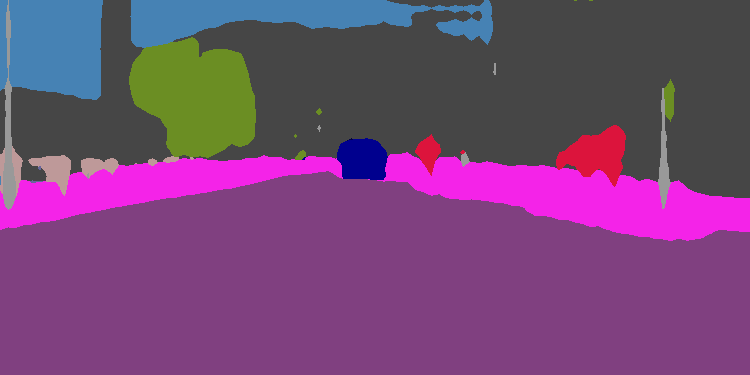} \\ 
   
&& & \includegraphics[width=\imgsize , valign=c]{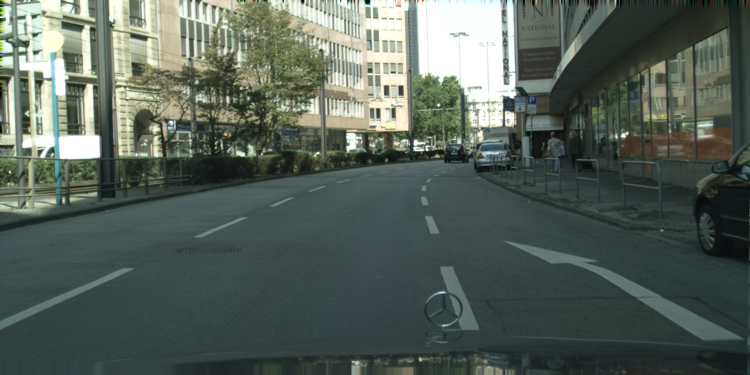} &
   \includegraphics[width=\imgsize , valign=c]{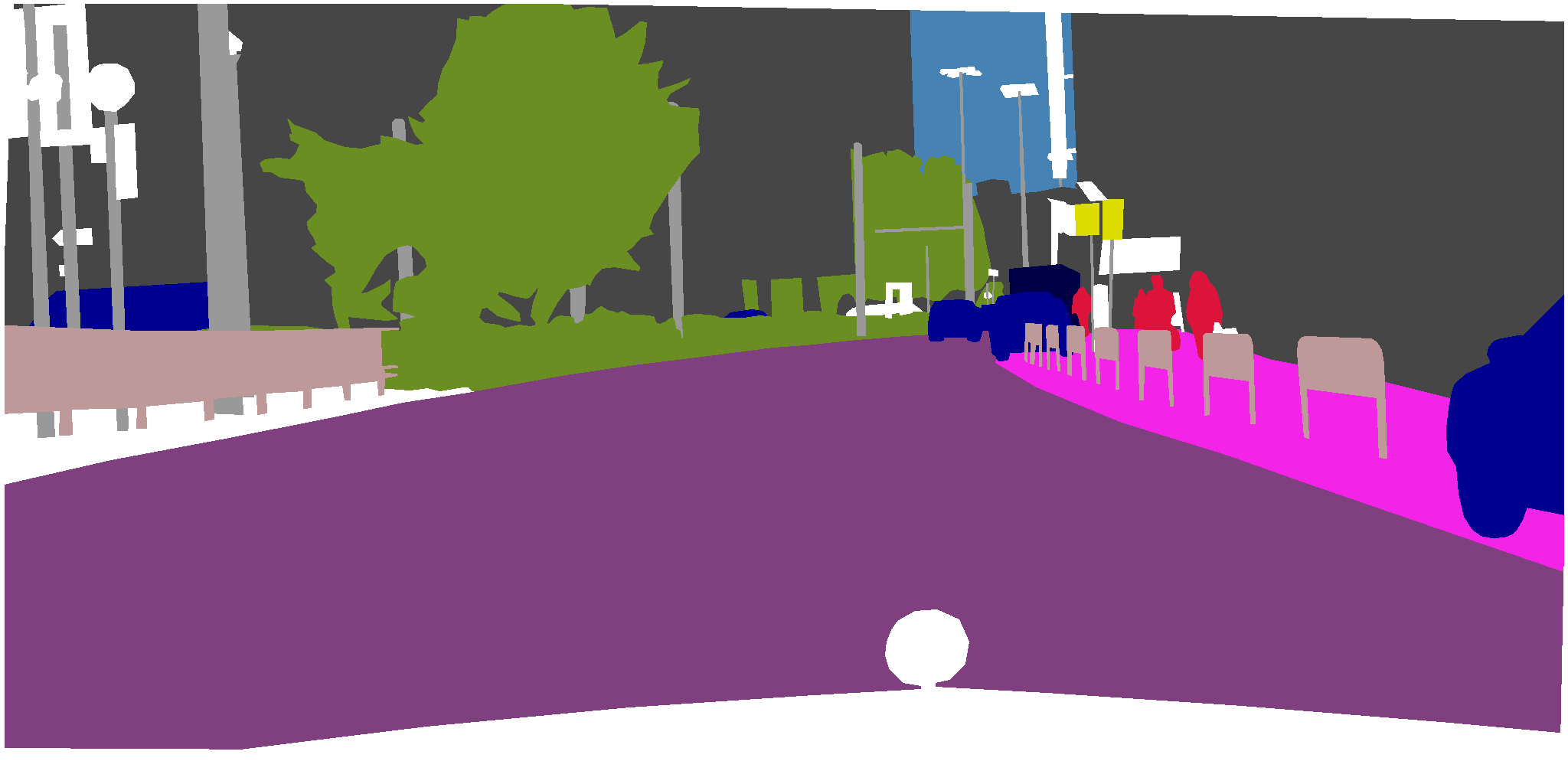} & 
   \includegraphics[width=\imgsize , valign=c]{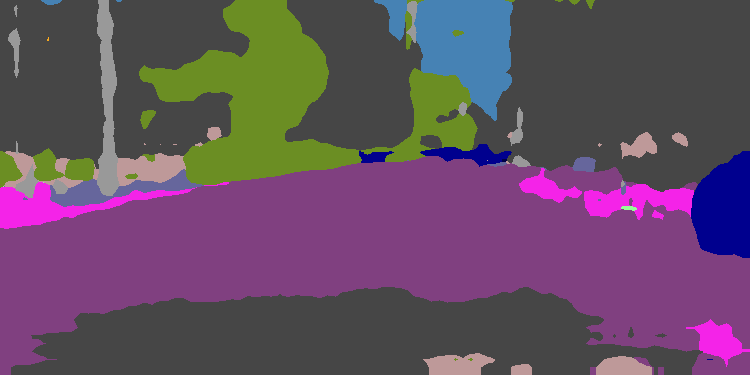} & 
   \includegraphics[width=\imgsize , valign=c]{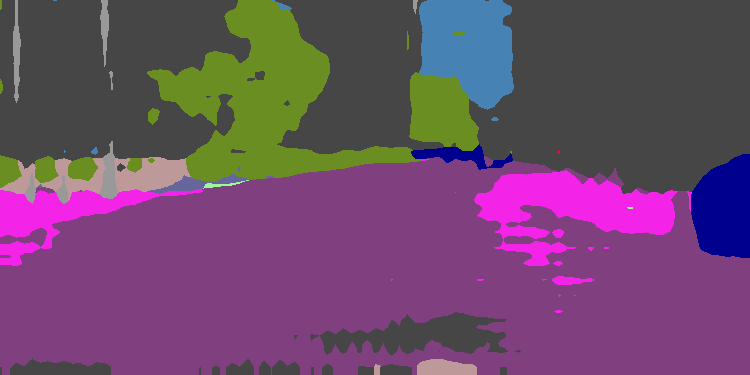} & 
   \includegraphics[width=\imgsize , valign=c]{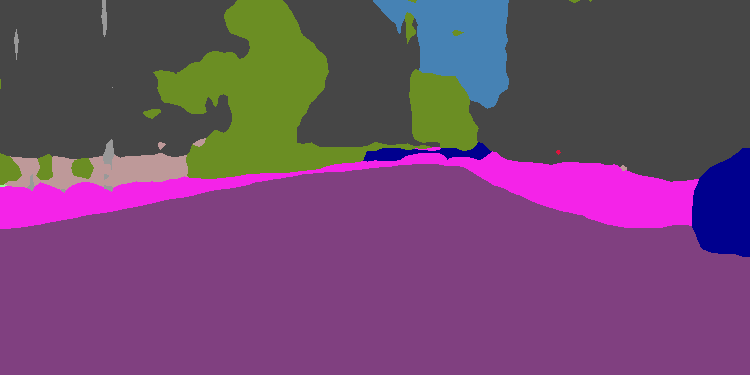}  \\  
   
%     && & \includegraphics[width=\imgsize]{images/DA_GTA-Cityscapes/rgb/frankfurt_000000_006589_leftImg8bit.png} &
%   \includegraphics[width=\imgsize]{images/DA_GTA-Cityscapes/groundtruth/frankfurt_000000_006589_gtFine_color.png} & 
%   \includegraphics[width=\imgsize]{images/DA_GTA-Cityscapes/supervised/frankfurt_000000_006589_leftImg8bit.png} & 
%   \includegraphics[width=\imgsize]{images/DA_GTA-Cityscapes/hung_et_al/frankfurt_000000_006589_leftImg8bit.png} & 
%   \includegraphics[width=\imgsize]{images/DA_GTA-Cityscapes/biasetton_et_al/frankfurt_000000_006589_leftImg8bit.png} & 
%   \includegraphics[width=\imgsize]{images/DA_GTA-Cityscapes/michieli_et_al/frankfurt_000000_006589_leftImg8bit.png} \\

   \cline{3-3}
  
   && \multirow{2}{*}{\rotatebox{90}{From SYNTHIA\hspace{-1ex}}} & 
   \includegraphics[width=\imgsize , valign=c]{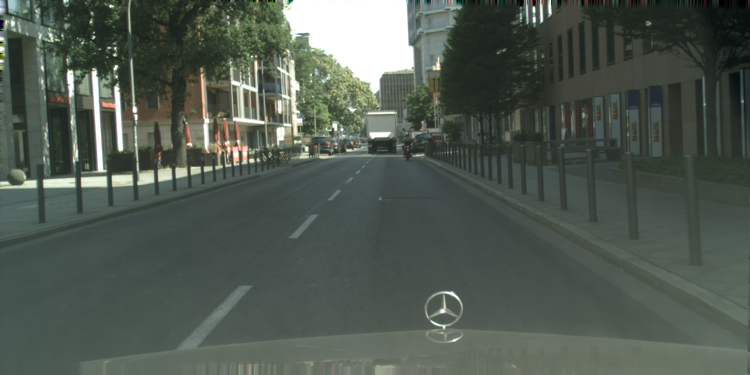} &
  \includegraphics[width=\imgsize , valign=c]{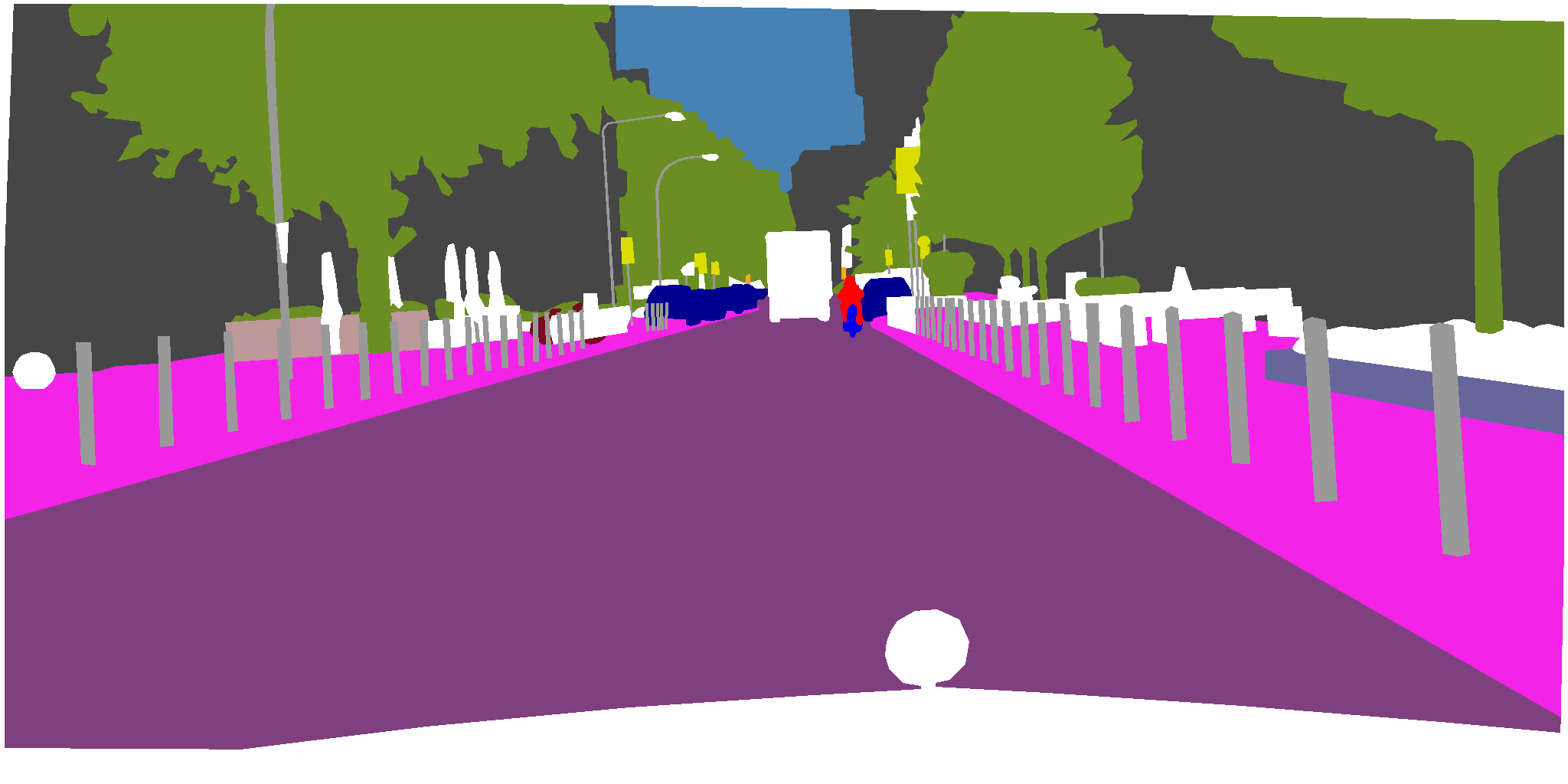} & 
  \includegraphics[width=\imgsize , valign=c]{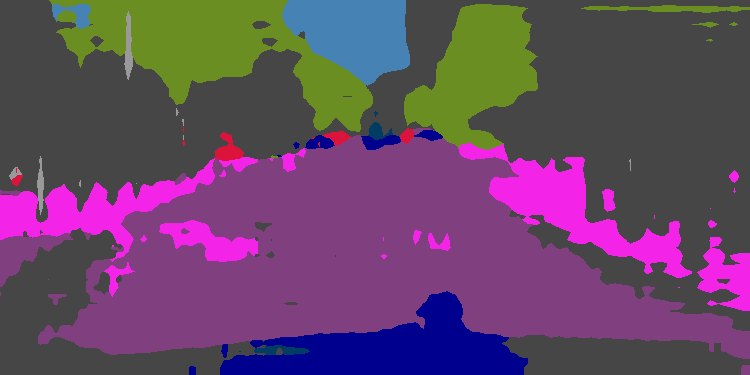} &  
  \includegraphics[width=\imgsize , valign=c]{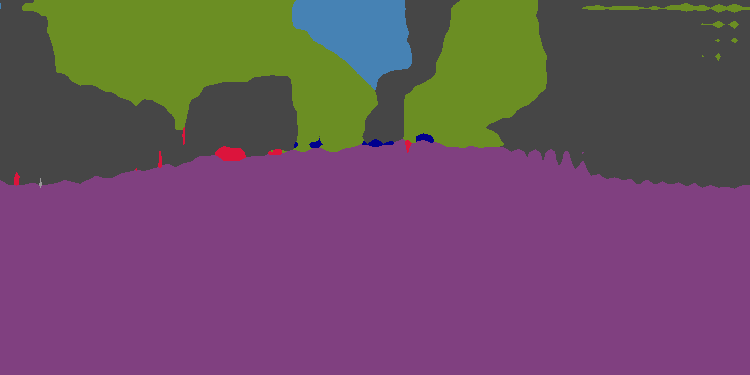} & 
  \includegraphics[width=\imgsize , valign=c]{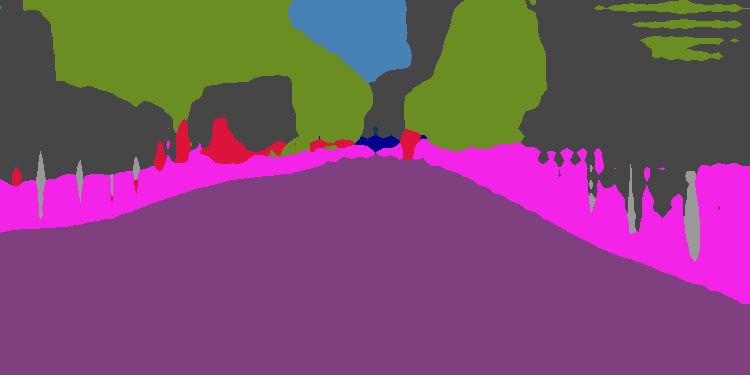}\\
 
  && & \includegraphics[width=\imgsize , valign=c]{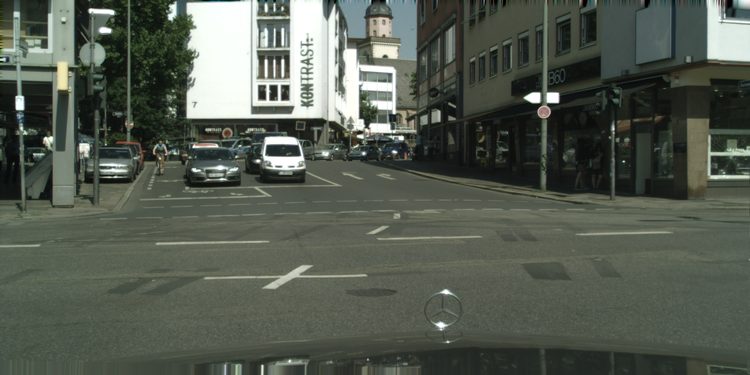} &
  \includegraphics[width=\imgsize , valign=c]{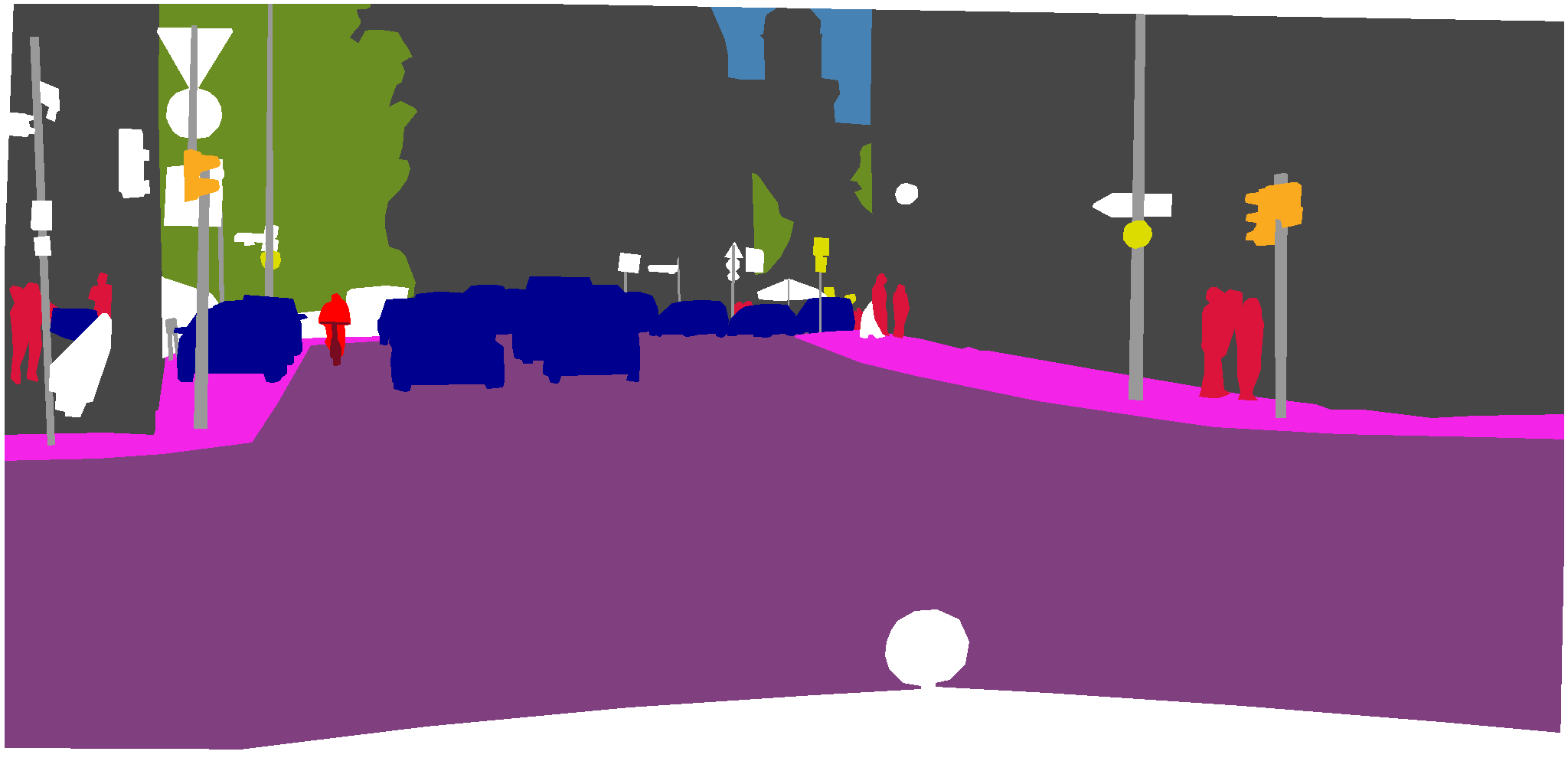} & 
  \includegraphics[width=\imgsize , valign=c]{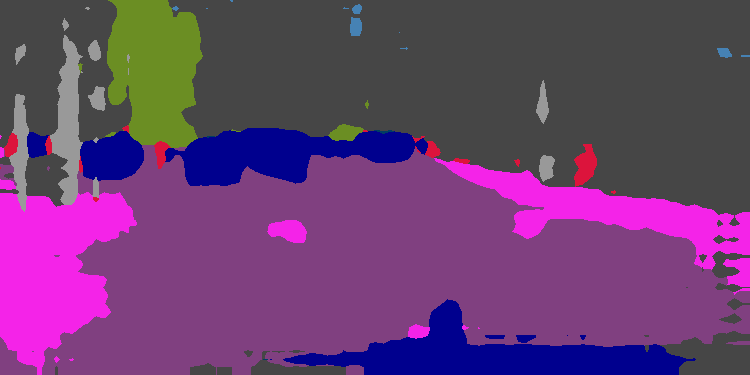} & 
  \includegraphics[width=\imgsize , valign=c]{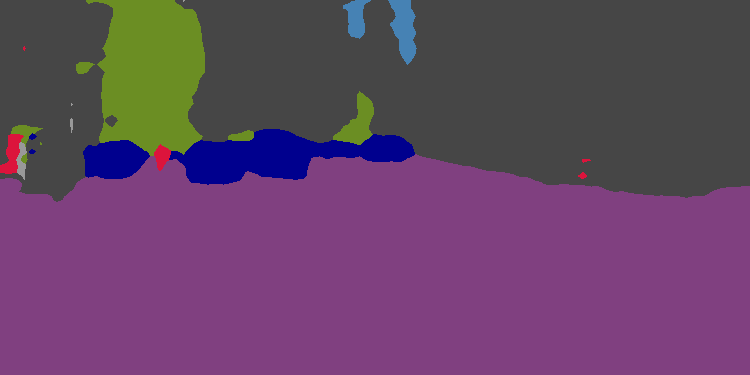} & 
  \includegraphics[width=\imgsize , valign=c]{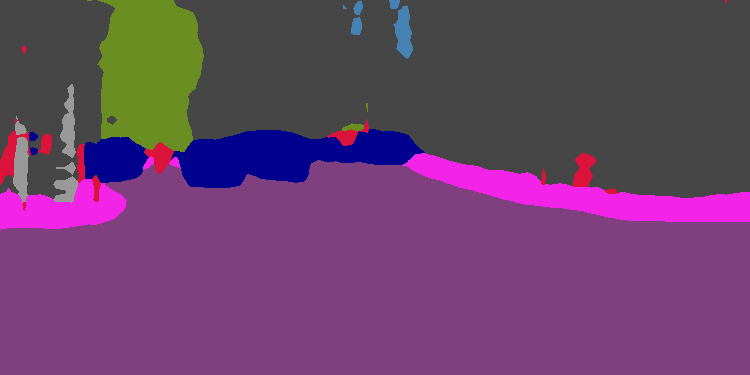} \\
  
%  && & \includegraphics[width=\imgsize]{images/DA_GTA-Cityscapes/rgb/frankfurt_000000_006589_leftImg8bit.png} &
%  \includegraphics[width=\imgsize]{images/DA_SYNTHIA_Cityscapes/groundtruth/frankfurt_000000_006589_gtFine_color.png} & 
%  \includegraphics[width=\imgsize]{images/DA_SYNTHIA_Cityscapes/supervised/frankfurt_000000_006589_leftImg8bit.png} & 
%  \includegraphics[width=\imgsize]{images/DA_SYNTHIA_Cityscapes/hung_et_al/frankfurt_000000_006589_leftImg8bit.png} & 
%  \includegraphics[width=\imgsize]{images/DA_SYNTHIA_Cityscapes/biasetton_et_al/frankfurt_000000_006589_leftImg8bit.png} & 
%  \includegraphics[width=\imgsize]{images/DA_SYNTHIA_Cityscapes/michieli_et_al/frankfurt_000000_006589_leftImg8bit.png} \\

  \cline{2-3}
  
\end{tabular}
\end{subfigure}
%\label{fig:GTA_Cityscapes}

\vspace{0.15cm}
\setlength{\tabcolsep}{1pt}
\centering
\begin{subfigure}[htbp]{2\textwidth}
\begin{tabular}{c|c|c|ccccc}

\cline{2-3}
    \multirow{6}{*}{\vspace{-18ex}b)} & \multirow{6}{*}{\rotatebox{90}{\hspace{-15ex}To Mapillary}} & \multirow{3}{*}{\rotatebox{90}{From GTA5\hspace{+1ex}}} &
   \includegraphics[width=\imgsize , valign=c]{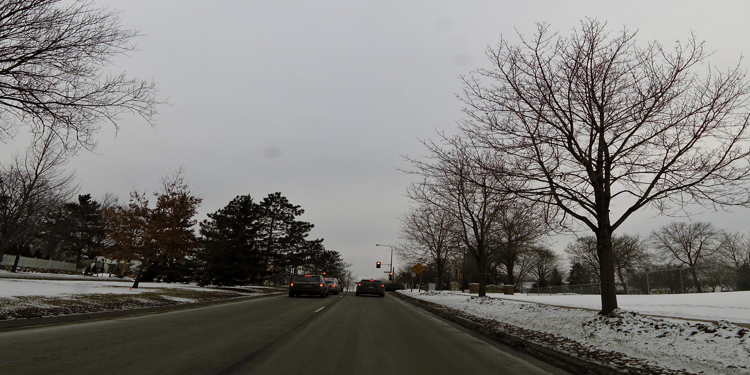} &
   \includegraphics[width=\imgsize , valign=c]{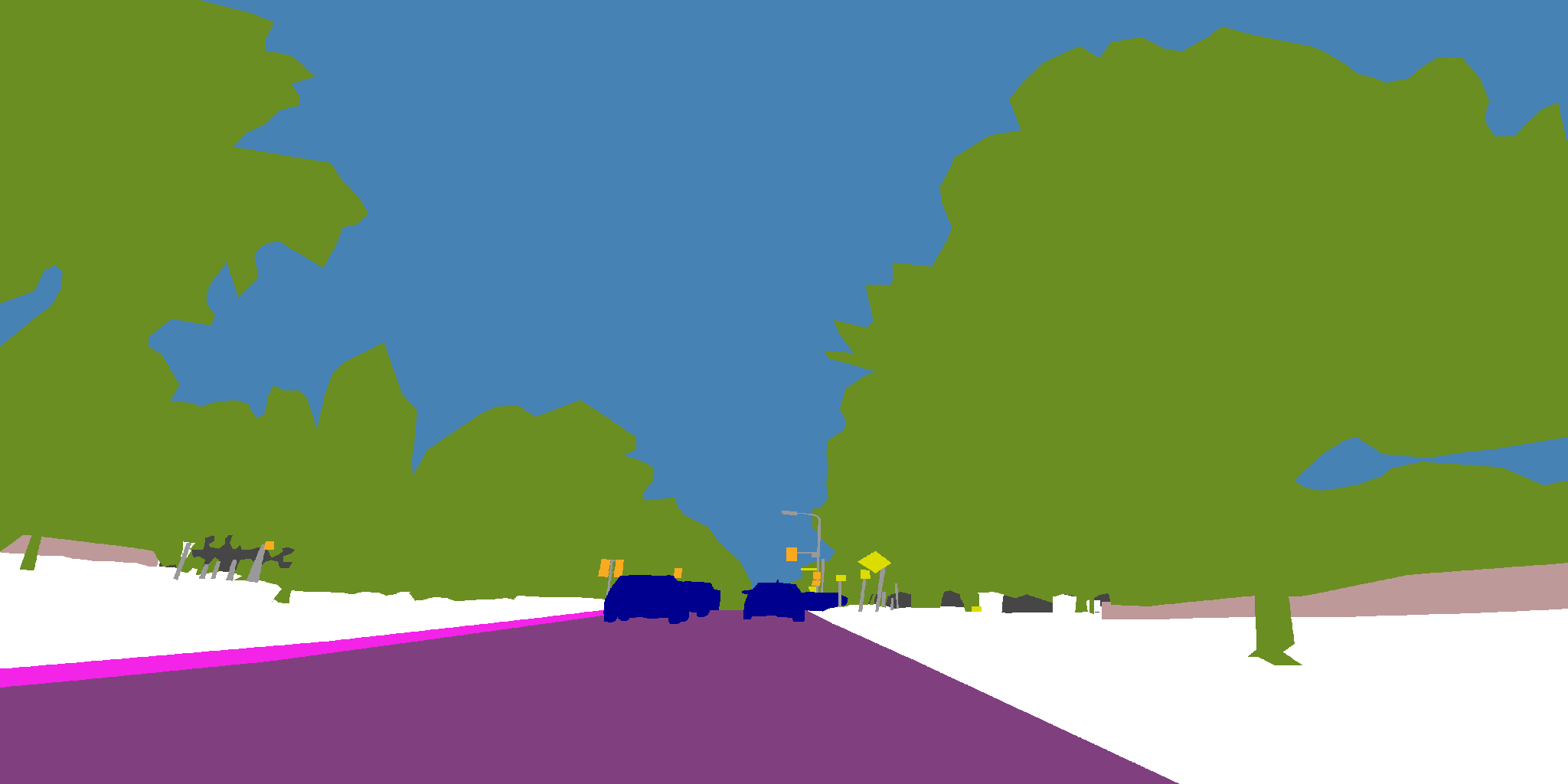} & 
   \includegraphics[width=\imgsize , valign=c]{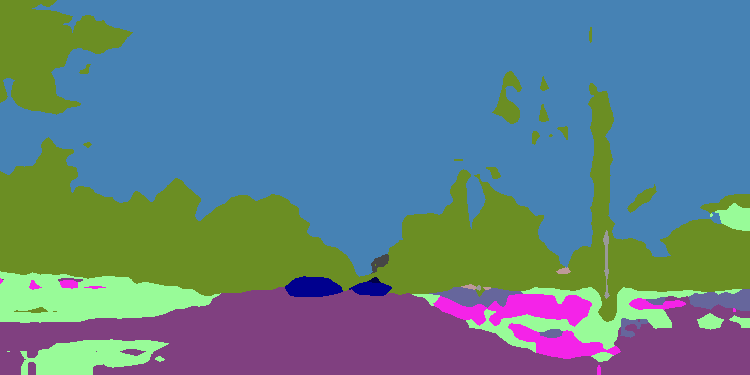} & 
   \includegraphics[width=\imgsize , valign=c]{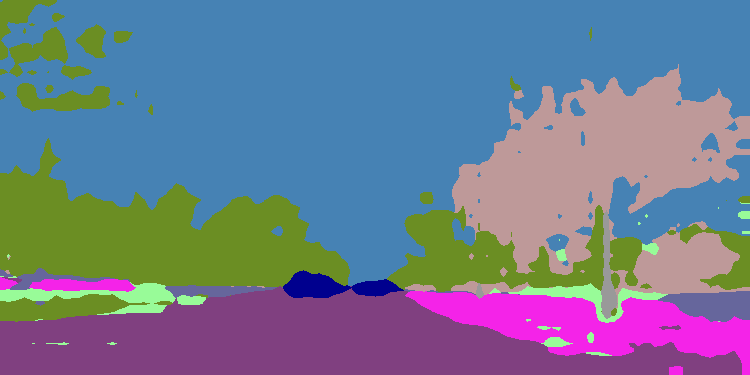} & 
   \includegraphics[width=\imgsize , valign=c]{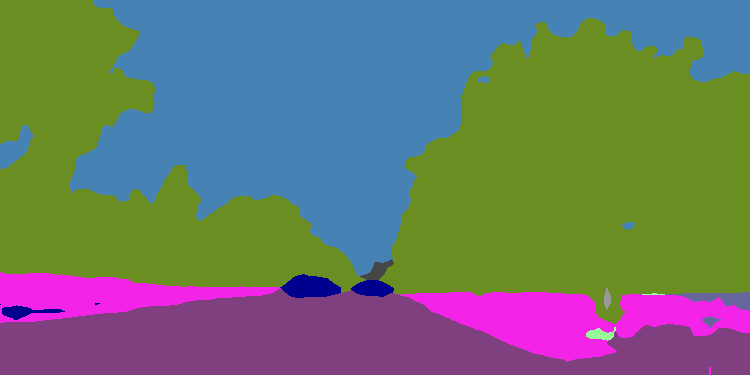} \\
  
  && & \includegraphics[width=\imgsize , valign=c]{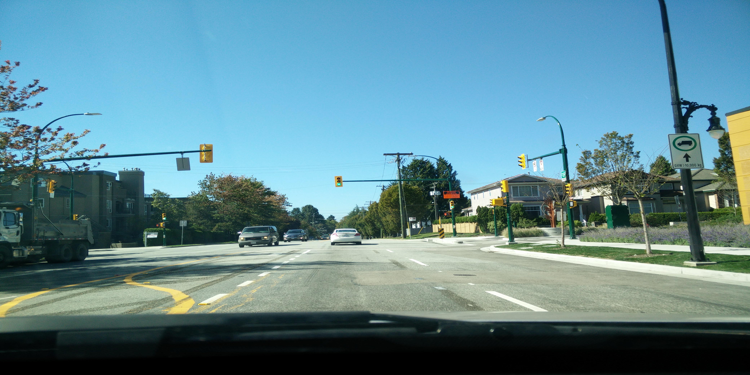} &
   \includegraphics[width=\imgsize , valign=c]{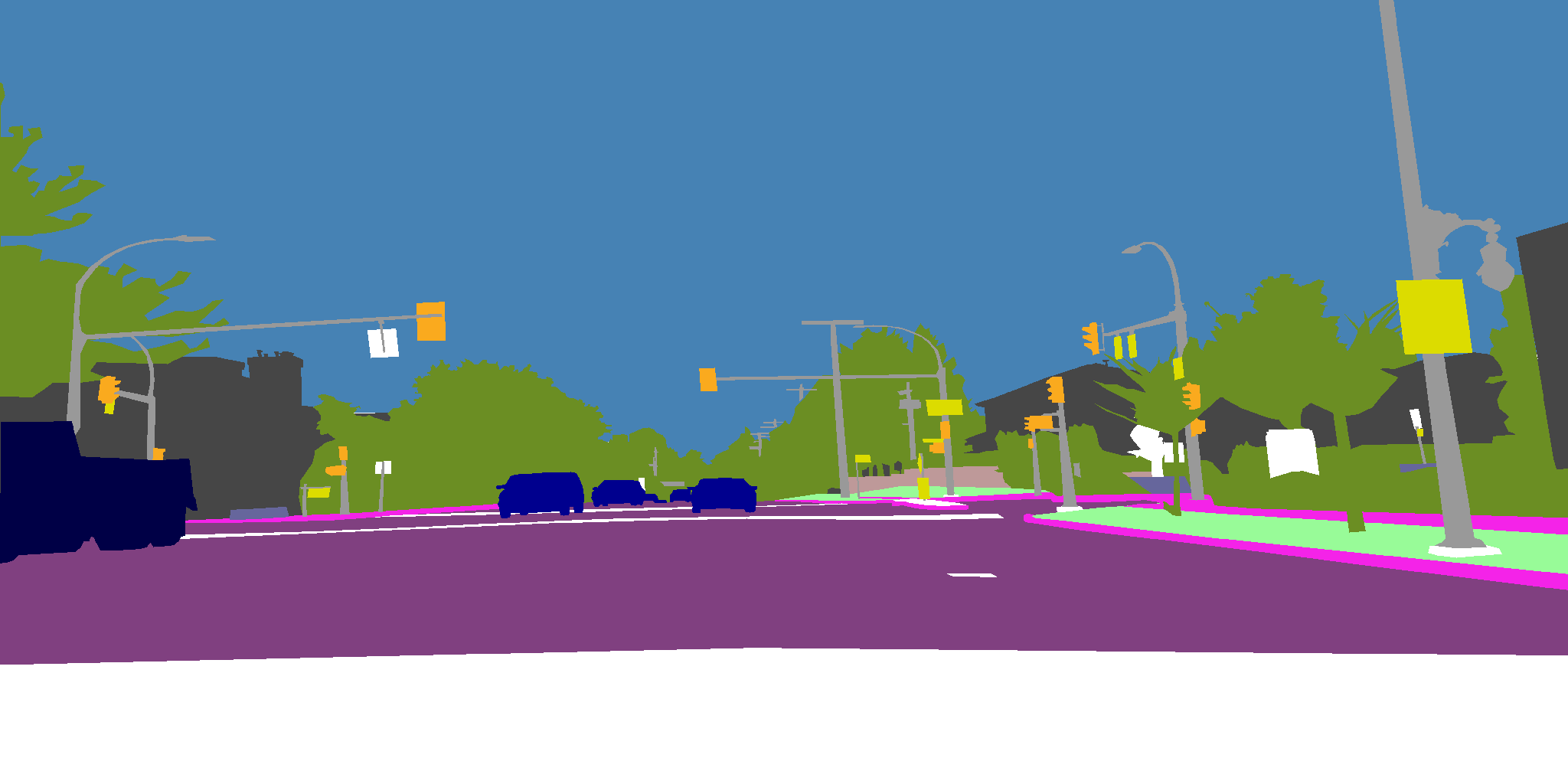} & 
   \includegraphics[width=\imgsize , valign=c]{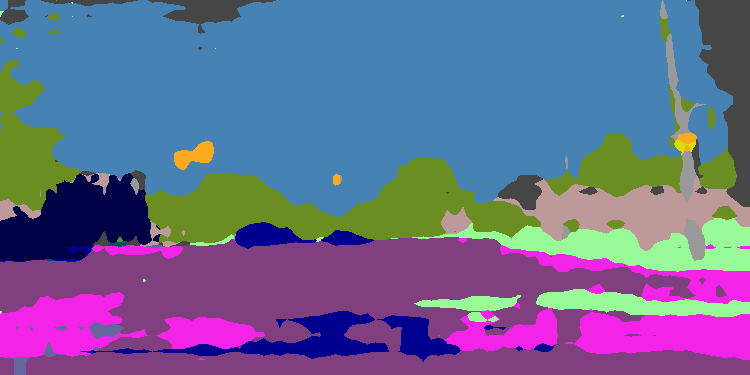} & 
   \includegraphics[width=\imgsize , valign=c]{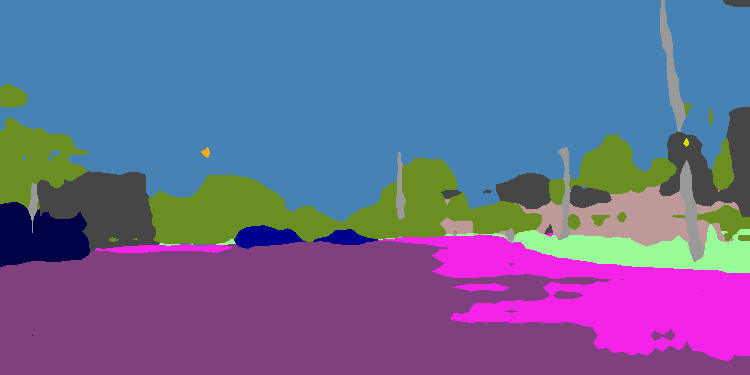} & 
   \includegraphics[width=\imgsize , valign=c]{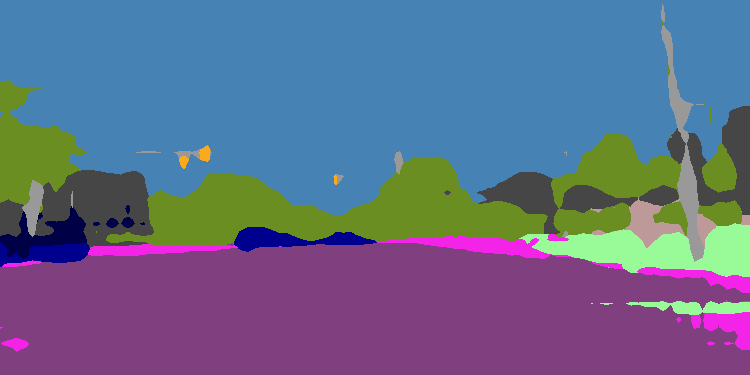} \\

%   && & \includegraphics[width=\imgsize]{images/DA_GTA-Mapillary/rgb/6hYSY9s_Z0wv_qHQc9fuJg.jpg} &
%   \includegraphics[width=\imgsize]{images/DA_GTA-Mapillary/groundtruth/6hYSY9s_Z0wv_qHQc9fuJg.png} & 
%   \includegraphics[width=\imgsize]{images/DA_GTA-Mapillary/supervised/6hYSY9s_Z0wv_qHQc9fuJg.png} & 
%   \includegraphics[width=\imgsize]{images/DA_GTA-Mapillary/hung_et_al/6hYSY9s_Z0wv_qHQc9fuJg.png} & 
%   \includegraphics[width=\imgsize]{images/DA_GTA-Mapillary/biasetton_et_al/6hYSY9s_Z0wv_qHQc9fuJg.png} & 
%   \includegraphics[width=\imgsize]{images/DA_GTA-Mapillary/michieli_et_al/6hYSY9s_Z0wv_qHQc9fuJg.png} \\
   \cline{3-3}
  
   && \multirow{3}{*}{\rotatebox{90}{From SYNTHIA\hspace{-1ex}}} & 
   \includegraphics[width=\imgsize , valign=c]{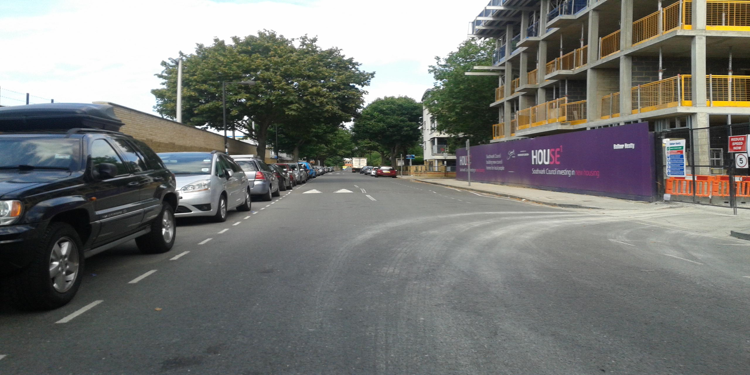} &
  \includegraphics[width=\imgsize , valign=c]{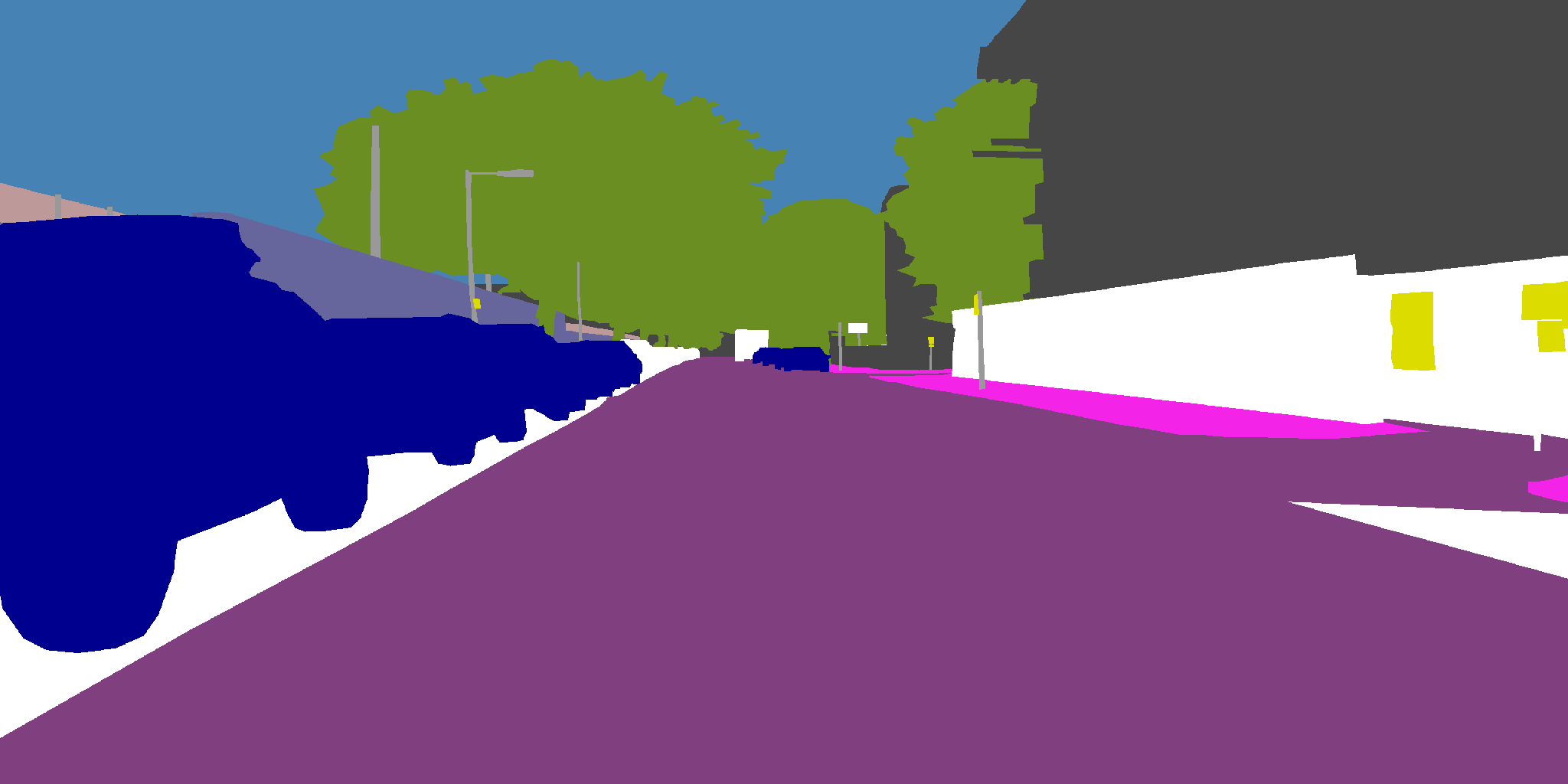} & 
  \includegraphics[width=\imgsize , valign=c]{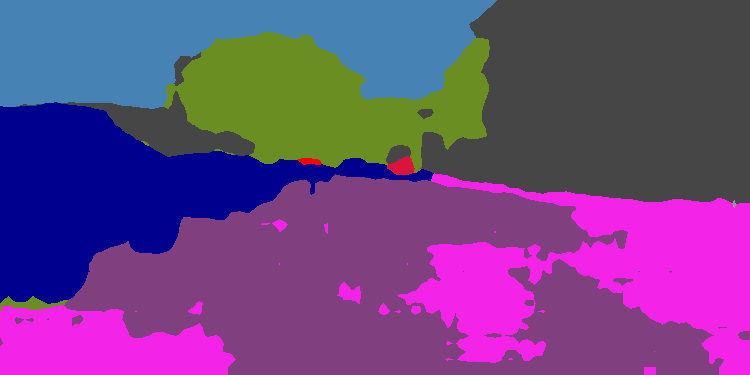} &  
  \includegraphics[width=\imgsize , valign=c]{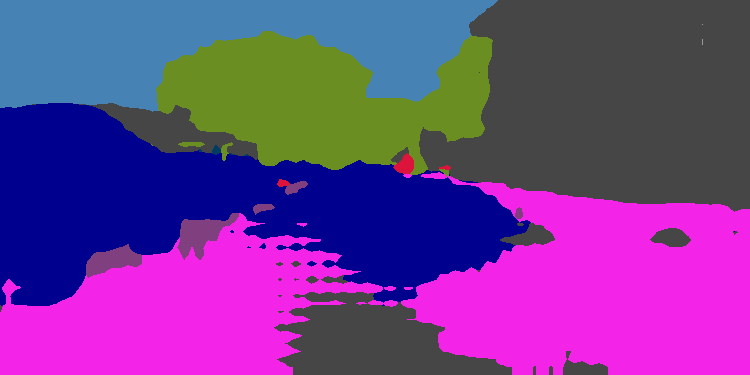} & 
  \includegraphics[width=\imgsize , valign=c]{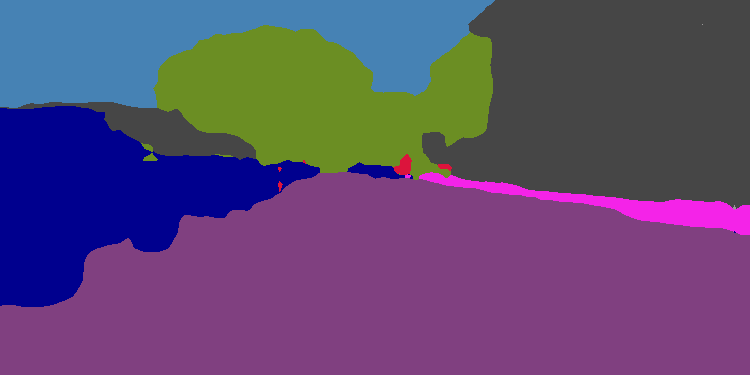} \\

 && & \includegraphics[width=\imgsize , valign=c]{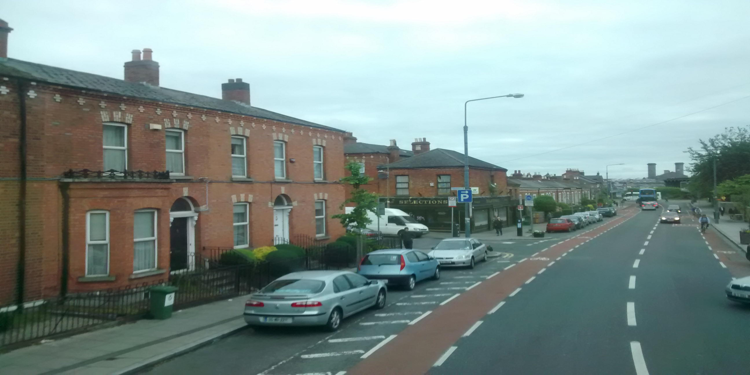} &
  \includegraphics[width=\imgsize , valign=c]{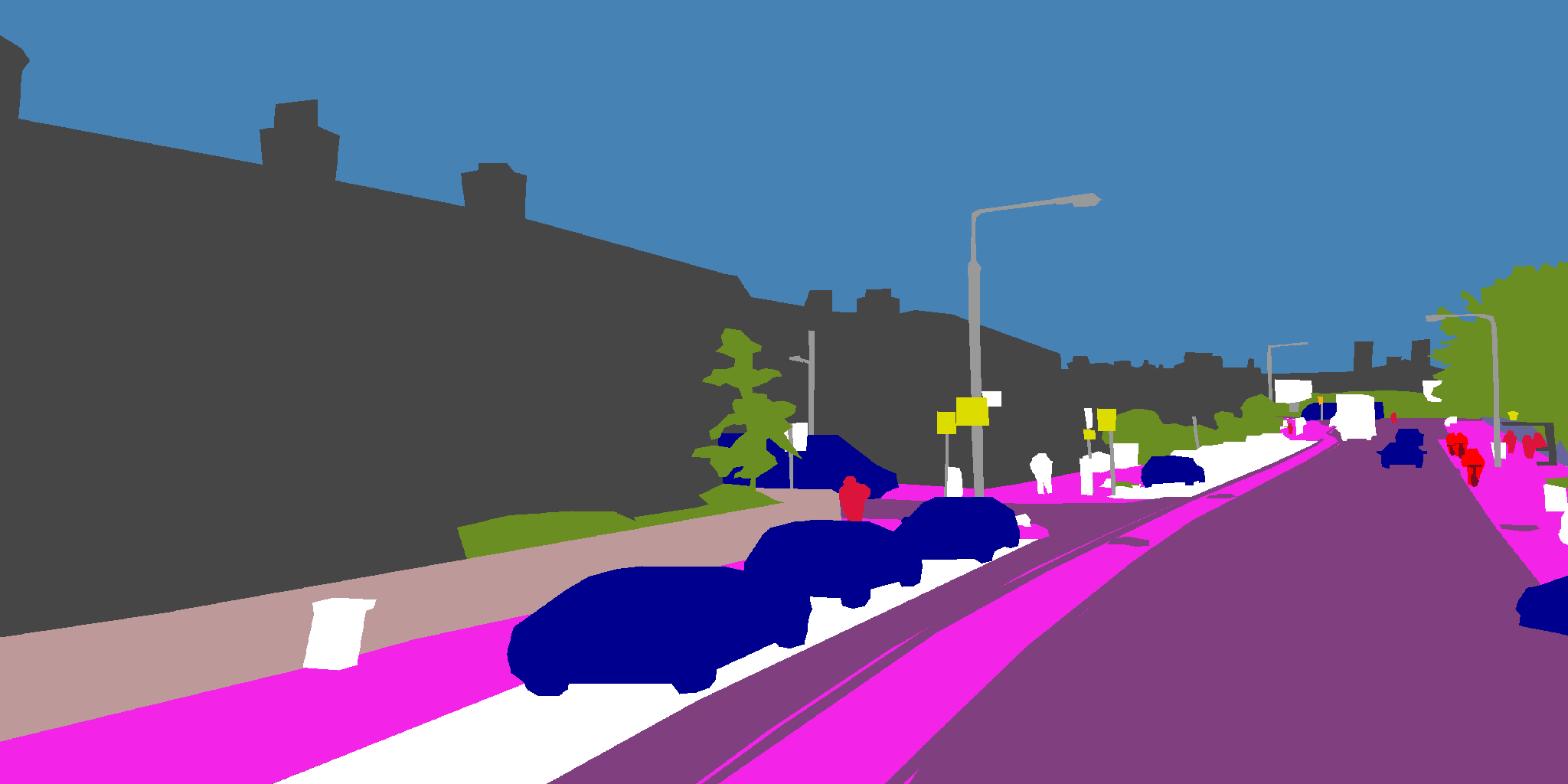} & 
  \includegraphics[width=\imgsize , valign=c]{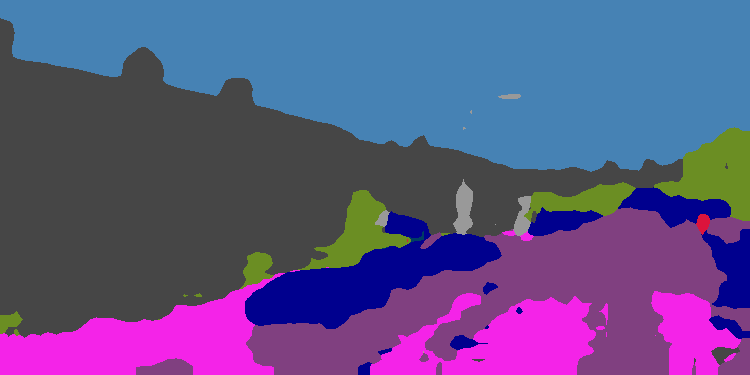} & 
  \includegraphics[width=\imgsize , valign=c]{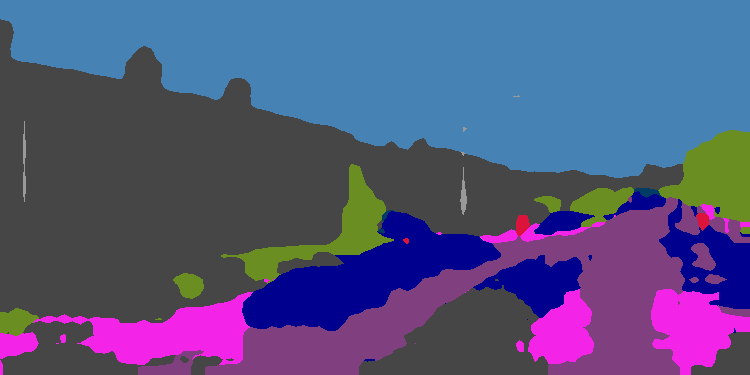} & 
  \includegraphics[width=\imgsize , valign=c]{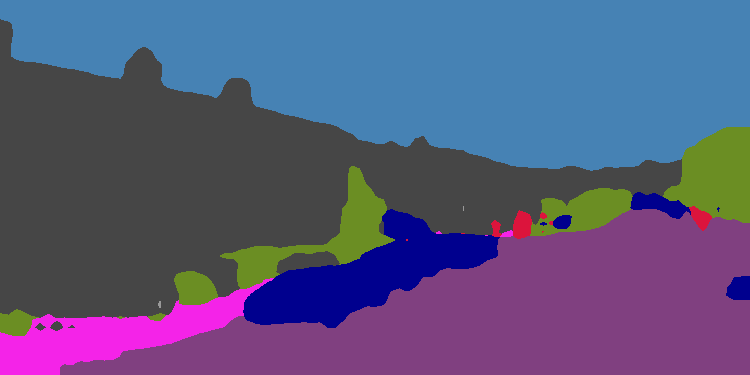} \\

%  & && \includegraphics[width=\imgsize]{images/DA_SYNTHIA-Mapillary/rgb/6hYSY9s_Z0wv_qHQc9fuJg.jpg} &
%  \includegraphics[width=\imgsize]{images/DA_SYNTHIA-Mapillary/groundtruth/6hYSY9s_Z0wv_qHQc9fuJg.png} & 
%  \includegraphics[width=\imgsize]{images/DA_SYNTHIA-Mapillary/supervised/6hYSY9s_Z0wv_qHQc9fuJg.png} & 
%  \includegraphics[width=\imgsize]{images/DA_SYNTHIA-Mapillary/hung_et_al/6hYSY9s_Z0wv_qHQc9fuJg.png} & 
%  \includegraphics[width=\imgsize]{images/DA_SYNTHIA-Mapillary/biasetton_et_al/6hYSY9s_Z0wv_qHQc9fuJg.png} & 
%  \includegraphics[width=\imgsize]{images/DA_SYNTHIA-Mapillary/michieli_et_al/6hYSY9s_Z0wv_qHQc9fuJg.png} \\
 \cline{2-3}
\multicolumn{3}{c}{} & image & annotation & supervised ($\mathcal{L}_{G,1}$) & Biasetton et al. \cite{biasetton2019} & ours ($\mathcal{L}_{full}$)
\end{tabular}
\end{subfigure}
\caption{Semantic segmentation of some sample scenes extracted from the Cityscapes (a) and Mapillary (b) validation datasets. The first group of four rows is related to the Cityscapes dataset, the last four to the Mapillary dataset. For each group, the first two rows are related to the experiments in which the GTA5 dataset is used as source. The last two rows are related to the case in which the SYNTHIA dataset is used as source (\textit{best viewed in colors}).}
\label{fig:qual_res}
\end{figure*}
\subsection{Evaluation on the Mapillary dataset}

We evaluate our approach also on the Mapillary dataset.
We start by using the GTA5 dataset for the supervised training as before: the results are shown in Table~\ref{tab:quantitative}c). With no adaptation to the Mapillary dataset the network achieves a mIoU of $37.8\%$. Thanks to the multiple domain discriminators and to the adaptive self-training techniques our framework is able to reach a mIoU of $41.9\%$, significantly outperforming all the compared methodologies. We can notice that the improvement with respect to the baseline approach is consistently distributed among the semantic classes and it is particularly evident on the \textit{road} or \textit{building} ones.
Qualitative results are shown in the first two rows of Fig.~\ref{fig:qual_res}b), where we can verify that most of the noise present in the supervised training and in \cite{biasetton2019} is filtered out by the proposed framework. In particular, the \textit{vegetation} and \textit{sidewalk} categories highly benefit from the domain adaptation with class-wise and time-variable %step-dependent, step-variable, sample-dependent
confidence threshold selection.

Furthermore, we can appreciate that, also on the Mapillary dataset, the accuracy is lower when the SYNTHIA dataset is used for supervised training, leading to a mIoU of $31.3\%$ only. As already noticed from the evaluations on Cityscapes, some classes (i.e., \textit{road}, \textit{sidewalk} and \textit{building}) have a low accuracy due to the poor texture representation and vastly profit by the adaptation to the target domain. 
The complete framework increases the final mIoU to  $34.9\%$ with an improvement of $3.8\%$, consistent with the previous experiments made across different datasets. Remarkable are the percentage gains in the aforementioned classes: for example,  the \textit{road} class more than doubles its accuracy and \textit{sidewalk}'s IoU grows by $12.2\%$. 
The visual results are reported in the last two rows of Fig.~\ref{fig:qual_res}b).  Here, for example, we can appreciate that the proposed approach is the only one to achieve an accurate and reliable recognition of \textit{road} and \textit{cars} classes on the shown images, which confirms our previous analysis.

\subsection{Ablation Study}
\label{sec:ablation}

In this section we present an accurate investigation of the effectiveness of the various modules of the proposed framework. For this study we consider the performances on the Mapillary dataset when adapting from GTA5.
We start by evaluating the individual impact of each module: the performance analysis is shown in Table~\ref{tab:ablation}. Let us recall that the baseline architecture, i.e., the Deeplab-v2 network trained on synthetic data only, achieves a mIoU of $37.8\%$ on real data. Then, we analyze the remaining modules by removing one component at a time (row $2$ to $6$). 
We can appreciate that all the components bring a significant contribution to the final mIoU, which in the full version of the approach where all of them are enabled is $41.9\%$. 
The impact of $\mathcal{L}_{G,1}^s$, $\mathcal{L}_{G,1}^t$ and $\mathcal{L}_{G,2}$ is 	clear by looking at Table~\ref{tab:ablation}:  without each of them the accuracy decreases with respect to the complete framework but remains higher than the source supervised case.
In particular, we performed a more detailed analysis of the self-training module. Having no self-training leads to $41.1\%$ of mIoU, while having self-training done on all pixels (without thresholding with  $\mathcal{T}_f$) or using a fixed confidence threshold (e.g., setting a  constant value of  $\mathcal{T}_f=0.2$ as in \cite{biasetton2019}) leads to $40.6\%$ and $40.9\%$, respectively. %While the accuracy of the latter is slightly higher than the former, as we filtered out some noisy confidence values, we argue that in the proposed framework,
These results show that self-training is not effective if the reliable pixels are not accurately selected, e.g., if performed immutable over the classes and the training steps. 
On the other side we found that self-training is effective with confidence thresholds variable over classes and over training time. %boosting the accuracy by almost $1\%$.

\begin{table}[tbp]
\setlength{\tabcolsep}{6pt}
\centering
\begin{tabular}{cccccc|cc}
$\mathcal{L}_{G,0}$ & $\mathcal{L}_{G,1}^s$ & $\mathcal{L}_{G,1}^t$ & $\mathcal{L}_{G,2}^t$ & $\mathcal{L}_{G,3}$ & $\mathcal{T}_{f}$ &  mIoU\\\hline
\checkmark & & & & & & 37.8 \\
\checkmark & & \checkmark & \checkmark & \checkmark & \checkmark & 39.9 \\
\checkmark & \checkmark & & \checkmark & \checkmark & \checkmark & 40.3 \\
\checkmark & \checkmark & \checkmark & & \checkmark & \checkmark & 40.7 \\
\checkmark & \checkmark & \checkmark & \checkmark & & & 41.1 \\
\checkmark & \checkmark & \checkmark & \checkmark & \checkmark & & 40.6 \\ 
\checkmark & \checkmark & \checkmark & \checkmark & \checkmark & fix $0.2$& 40.9 \\
\checkmark & \checkmark & \checkmark & \checkmark & \checkmark & \checkmark & \textbf{41.9} \\
\end{tabular}
\caption{Ablation results on the Mapillary validation set adapting from GTA5.}
\label{tab:ablation}
\end{table}

\begin{figure}[tbp]
\centering
\includegraphics[width=\linewidth]{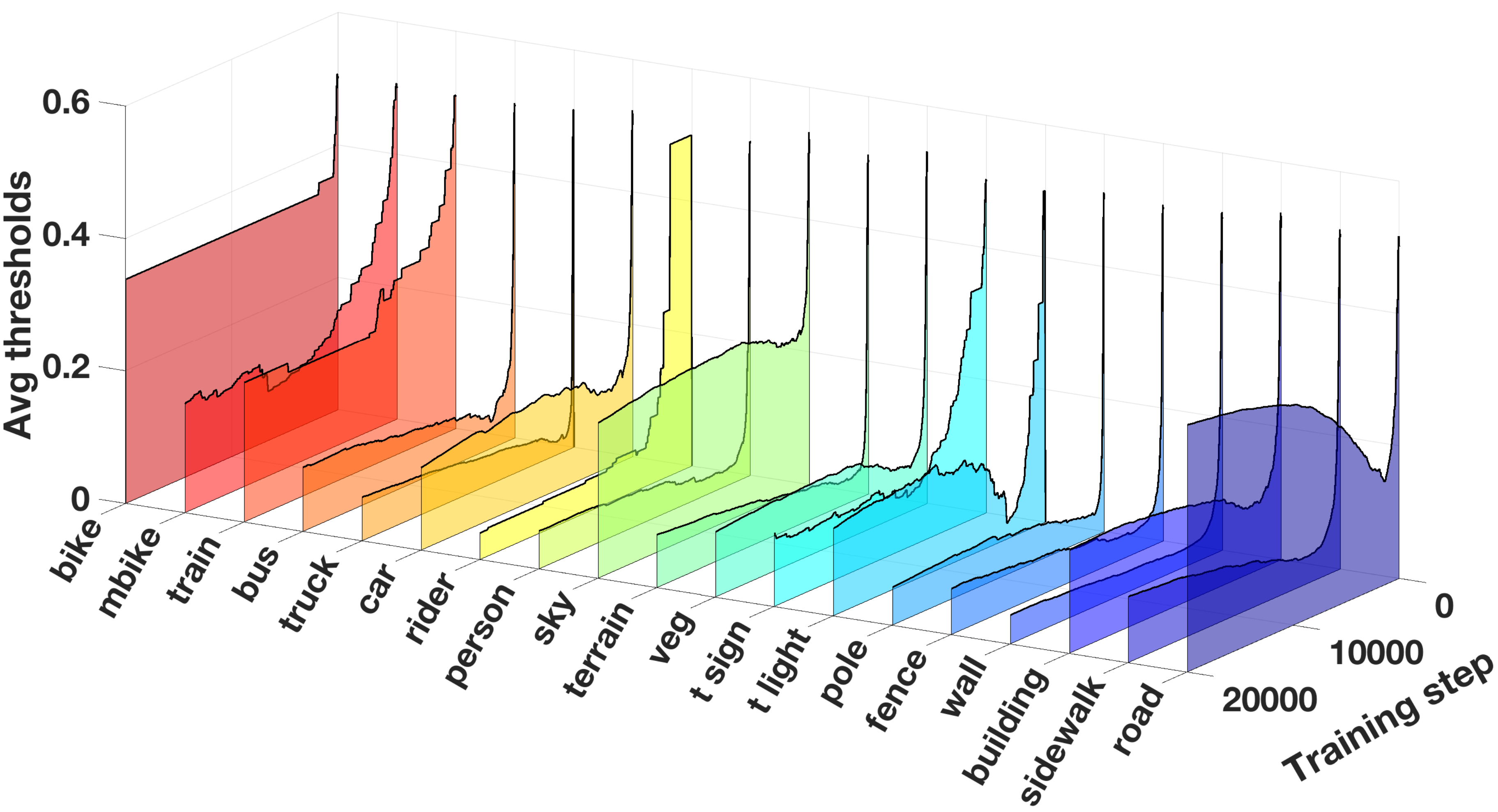}
%\caption{Average per-class confidence thresholds for different classes and at different training steps on the Mapillary dataset when adapting from GTA5 (\textit{best viewed in colors}).}
\caption{Time average over the initial to current step interval of per-class confidence thresholds for different classes and at different training steps on the Mapillary dataset when adapting from GTA5 (\textit{best viewed in colors}).}
\label{fig:plot_thresholds}
\end{figure}

We also analyzed the behavior of the per-class time-varying confidence values: they are shown in Fig.~\ref{fig:plot_thresholds} for the GTA5 to Mapillary scenario. Here, we can appreciate how the confidence thresholds vary over the training time and typically converge to a value which may be significantly different among the various classes. While previous works \cite{biasetton2019,michieli2020adversarial} fixed the threshold to $0.2$, here we can see how the desired value is variable and ranges between $0.04$ and $0.4$. Highly confident classes typically have high confidence threshold, in order to propagate back very reliable predictions through self-training as for instance \textit{road}, \textit{sky} and \textit{building}. However, notice that \textit{train} and \textit{bike} have high confidence threshold values only because they are predicted too rarely to be useful for confidence assessment (i.e., the thresholds are piece-wise constant functions), hence represent failure modes of the estimate. More challenging classes such as \textit{wall} and \textit{fence}, instead, are characterized by a much lower confidence value. The most important aspect, however, is to verify that our framework can adapt both to the properties of different classes and to the changes in the network behavior during the training procedure. Additionally, we can compute the mean threshold value averaged over all the classes at the end of the training phase and compare it with previous works \cite{biasetton2019,michieli2020adversarial}. The mean is  $0.13$ in the considered scenario and $0.19$ in the adaptation from GTA5 to Cityscapes,   consistently with the value of $0.20$ used in previous works \cite{biasetton2019,michieli2020adversarial}.

\newcommand{\dpiVal}{100}

\section{Conclusions}
\label{sec:conclusions}

In this paper, we investigated the unsupervised domain adaptation task from synthetic urban scenes to real world ones. Two different strategies have been used to exploit unlabeled data. Firstly, an adversarial learning framework based on a couple of fully convolutional discriminators has been employed to align the two domain distributions. 
Secondly, we designed a self-teaching strategy based on the assumption that unsupervised predictions labeled as highly confident by the discriminator are reliable. Thanks to class-aware and time-varying confidence thresholds our framework can adapt to different classes and different stages of learning of the semantic segmentation network.
We performed several experiments on two real datasets starting from two synthetic ones, thus proving the effectiveness and robustness of the proposed approach across different scenarios.
In the future, we plan to enrich the proposed framework by enforcing the alignment of the features spaces produced from the two domains and to test it in different contexts than the road scenario.

\bibliographystyle{IEEEtran}

\bibliography{strings_reduced,refs}

\end{document}